\documentclass[twoside]{article}

\usepackage[accepted]{aistats2024}
\usepackage[round]{natbib}

\usepackage{microtype}
\usepackage{graphicx}
\usepackage{subfigure}
\usepackage{booktabs} 
\usepackage{multirow}
\usepackage{chemfig}

\usepackage[utf8]{inputenc} 
\usepackage[T1]{fontenc}    
\usepackage{hyperref}       
\usepackage{url}            
\usepackage{booktabs}       
\usepackage{amsfonts}       
\usepackage{nicefrac}       
\usepackage{microtype}      
\usepackage{xcolor}         

\usepackage{amsmath}
\usepackage{amssymb}
\usepackage{mathtools}
\usepackage{amsthm}
\usepackage{bbm}
\usepackage{multirow}
\usepackage{lipsum}
\usepackage[frak = euler]{mathalpha}

\usepackage[capitalize,noabbrev]{cleveref}
\usepackage{enumerate}

\begin{document}

%

%

\twocolumn[

\aistatstitle{Hierarchically Coherent Multivariate Mixture Networks}

\aistatsauthor{ Kin G. Olivares* \And David Luo* \And  Cristian Challu \And Stefania La Vattiata}
\aistatsauthor{ Max Mergenthaler \And Artur Dubrawski}

\aistatsaddress{ Auton Lab \And  Carnegie Mellon University \And Nixtla }]

\theoremstyle{plain}
\newtheorem{theorem}{Theorem}[section]
\newtheorem{proposition}[theorem]{Proposition}
\newtheorem{lemma}[theorem]{Lemma}
\newtheorem{corollary}[theorem]{Corollary}
\theoremstyle{definition}
\newtheorem{definition}[theorem]{Definition}
\newtheorem{assumption}[theorem]{Assumption}
\theoremstyle{remark}
\newtheorem{remark}[theorem]{Remark}

\long\def\KO#1{{\color{olive}{[KO: #1]}\color{black}}}
\long\def\DL#1{{\color{magenta}{[DL: #1]}\color{black}}}
\long\def\CC#1{{\color{lime}{[CC: #1]}\color{black}}}
\long\def\SL#1{{\color{violet}{[SL: #1]}\color{black}}}
\long\def\MM#1{{\color{orange}{[MM: #1]}\color{black}}}
\long\def\AD#1{{\color{red}{[AD: #1]}\color{black}}}
\long\def\EDIT#1{{\color{black}{#1}\color{black}}}

\newcommand{\model}[1]{\texttt{#1}}
\newcommand{\ours}{\model{{HINT}}}
\newcommand{\ourstext}{Hierarchically Coherent Multivariate Mixture Network}
\newcommand{\TemporalNorm}{\model{TemporalNorm}}

\newcommand{\ARIMA}{\model{ARIMA}}
\newcommand{\Naive}{\model{Naive}}
\newcommand{\GMM}{\model{GMM}}
\newcommand{\MLP}{\model{MLP}}
\newcommand{\TFT}{\model{TFT}}
\newcommand{\TCN}{\model{TCN}}
\newcommand{\LSTM}{\model{LSTM}}
\newcommand{\ESRNN}{\model{ESRNN}}
\newcommand{\NHITS}{\model{NHITS}}
\newcommand{\NBEATS}{\model{NBEATS}}
\newcommand{\NBEATSx}{\model{NBEATSx}}
\newcommand{\PatchTST}{\model{PatchTST}}
\newcommand{\REVIN}{\model{REVIN}}
\newcommand{\Autoformer}{\model{Autoformer}}

\newcommand{\GMMtext}{Gaussian Mixture Mesh}

\newcommand{\DeepAR}{\model{DeepAR}}
\newcommand{\PMMCNN}{\model{PMMCNN}}
\newcommand{\SeqtoSeqC}{\model{Seq2SeqC}}
\newcommand{\MQForecaster}{\model{MQ-Forecaster}}

\newcommand{\Base}{\model{Base}}
\newcommand{\BottomUp}{\model{BottomUp}}
\newcommand{\TopDown}{\model{TopDown}}
\newcommand{\MiddleOut}{\model{MiddleOut}}
\newcommand{\MinTrace}{\model{MinTrace}}
\newcommand{\Comb}{\model{Comb}}
\newcommand{\ERM}{\model{ERM}}

\newcommand{\OTHER}{\model{OTHER}}
\newcommand{\PERMBU}{\model{PERMBU}}
\newcommand{\NORMALITY}{\model{NORMALITY}}
\newcommand{\BOOTSTRAP}{\model{BOOTSTRAP}}

\newcommand{\HIRED}{\model{HIRED}}
\newcommand{\HierEtoE}{\model{HierE2E}}
\newcommand{\SHARQ}{\model{SHARQ}}
\newcommand{\PROFHIT}{\model{PROFHIT}}
\newcommand{\TDProb}{\model{TDProb}}

\newcommand{\AutoARIMA}{\model{AutoARIMA}}
\newcommand{\StatsForecast}{\model{StatsForecast}}
\newcommand{\NeuralForecast}{\model{neuralforecast}}
\newcommand{\HierarchicalForecast}{\model{HierarchicalForecast}}

\newcommand{\ADAM}{\model{ADAM}}
\newcommand{\PyTorch}{\model{PyTorch}}
\newcommand{\HYPEROPT}{\model{HYPEROPT}}

\newcommand{\Smatrix}{\mathbf{S}_{[a,b][b]}}
\newcommand{\Amatrix}{\mathbf{A}_{[a][b]}}

\newcommand{\CRPSgainsLabour}{{8.2\%}}
\newcommand{\CRPSgainsTraffic}{{-11.3\%}}
\newcommand{\CRPSgainsTourismS}{{17.4\%}}
\newcommand{\CRPSgainsTourismL}{{10.9\%}}
\newcommand{\CRPSgainsWikitwo}{{15.5\%}}


\newcommand{\dataset}[1]{\texttt{#1}}
\newcommand{\Labour}{\dataset{Labour}}
\newcommand{\Tourism}{\dataset{Tourism}}
\newcommand{\TourismL}{\dataset{Tourism-L}}
\newcommand{\Traffic}{\dataset{Traffic}}
\newcommand{\Wikitwo}{\dataset{Wiki2}}
\newcommand{\Favorita}{\dataset{Favorita}}
\newcommand{\Mfive}{\dataset{M5}}

\newcommand\blambda{\boldsymbol \lambda}
\newcommand\btheta{\boldsymbol \theta}
\newcommand\bomega{\boldsymbol \omega}
\newcommand\bmu{\boldsymbol \mu}
\newcommand\bsigma{\boldsymbol \sigma}
\newcommand\bSigma{\boldsymbol \Sigma}


\begin{abstract}
   Large collections of time series data are often organized into hierarchies with different levels of aggregation; examples include product and geographical groupings. Probabilistic coherent forecasting is tasked to produce forecasts consistent across levels of aggregation. In this study, we propose to augment neural forecasting architectures with a coherent multivariate mixture output. We optimize the networks with a composite likelihood objective, allowing us to capture time series' relationships while maintaining high computational efficiency. Our approach demonstrates 13.2\% average accuracy improvements on most datasets compared to state-of-the-art baselines. We conduct ablation studies of the framework components and provide theoretical foundations for them. To assist related work, the code is available at this \href{https://anonymous.4open.science/r/HINT-FBA8/README.md}{\textcolor{blue}{http URL}}.
\end{abstract}


\section{Introduction} \label{section1:introduction}
Time series data is often organized into hierarchical structures spanning different levels of aggregation. Independently forecasting any series from such a hierarchy is unlikely to produce
 coherent results across the levels of the hierarchy, that is, forecasts that satisfy the aggregation constraints of the original
data. In recent years, new hierarchical reconciliation has become standard to achieve forecasts' coherence~\citep{babai2022hierarchical_review}, with applications to electricity generation~\citep{taieb2021hierarchical_electricity}, macroeconomics, and tourism~\citep{eckert2021hierarchical_exports, kourentzes2019hierarchical_tourism}.



\begin{figure}[t] 
    \centering
    \subfigure[\emph{Aggregation Constraints}]{
    \label{fig:hierarchical_example}
    \includegraphics[width=0.44\linewidth]
    {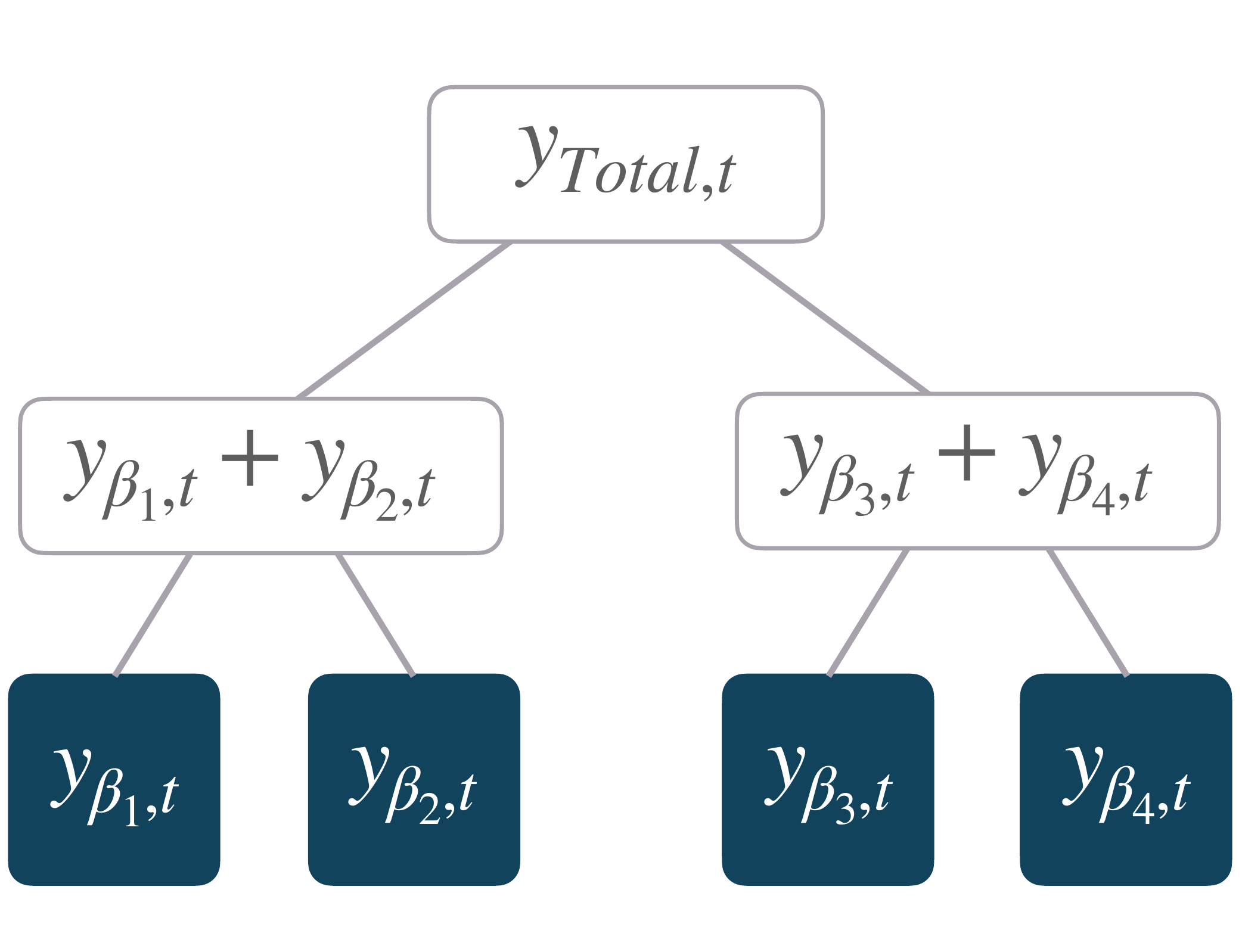}
    }
    \subfigure[\emph{Error Accumulation}]{
    \label{fig:performance}
    \includegraphics[width=0.44\linewidth]{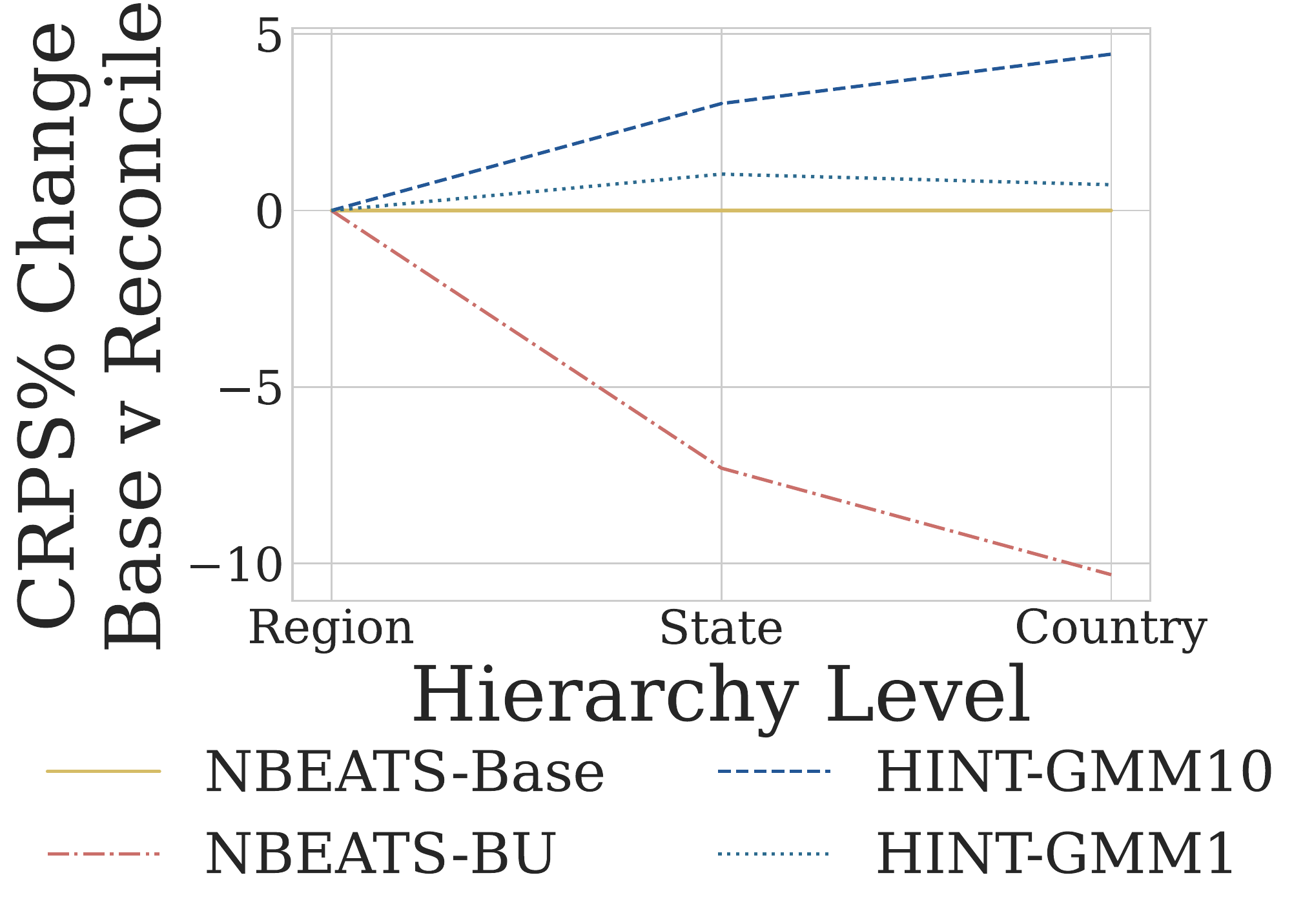} 
    }    
    \subfigure[\emph{Coherent Multivariate Mixture}]{
    \includegraphics[width=0.94\linewidth]{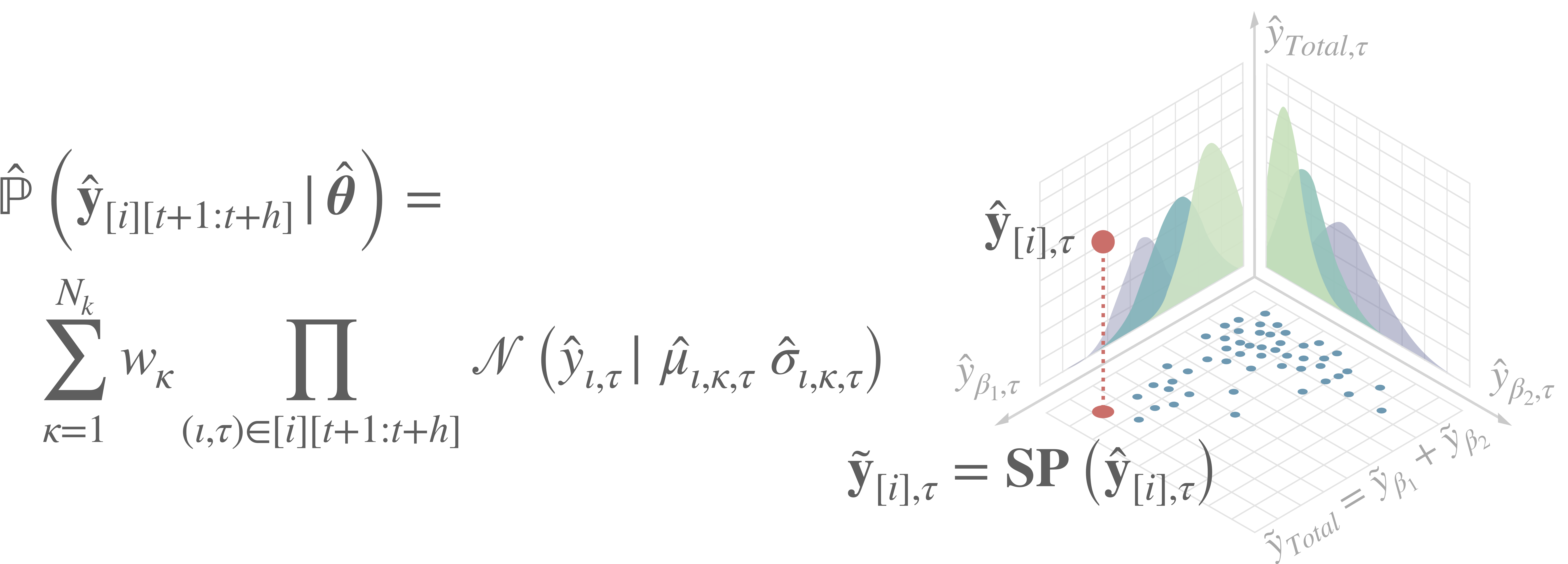} 
    }
    \caption{(a) Probabilistic coherent forecasting is a multivariate regression problem with aggregation constraints. (b) Summing disaggregated level's forecasts can accumulate errors. (c) Augmenting neural forecasting architectures with an 
    accurate and coherent multivariate mixture is a solution.
    }
    \label{fig:motivation}
\end{figure}

Previous neural hierarchical forecasting research is limited by restrictive probabilistic assumptions, weak coherence enforcement, poor scalability, and sometimes only modest accuracy improvements compared to statistical baselines. To address these limitations, we introduce the \emph{HIerarchically coherent multivariate mixture NeTworks} (\ours). Our contributions are: 
\begin{enumerate}[(i)]
    \itemsep0em 
    \item \textbf{Coherent Multivariate Mixture} for accurate and coherent forecasts, optimized with composite likelihood. Our approach captures series' relationships maintaining high computational efficiency.
    \item \textbf{Probabilistic Reversible Instance Normalization} to enhance forecast distributions against stark variations in scales of the series. Our approach normalizes inputs and reassembles probabilistic outputs via a global skip connection.
    \item \textbf{State-of-the-art results} 
     on relevant benchmark data: Australian labour, SF Bay Area traffic rates, Australian tourist visits, Wikipedia articles views.
\end{enumerate}

This paper is structured as follows. Section~\ref{section2:literature} reviews literature and introduces notation, Section~\ref{section3:methodology} describes the methodology, Section~\ref{section4:evaluation} analyzes our empirical results. Section~\ref{section5:discussion} discusses results and future research directions, while Section~\ref{section6:conclusion} concludes. Finally, in the Appendix, we study the effects of the individual components of \ours\  and its theoretical foundations.





\section{Related Work} \label{section2:literature}
\begin{figure*}[t] 
\centering
\includegraphics[width=0.92\linewidth]{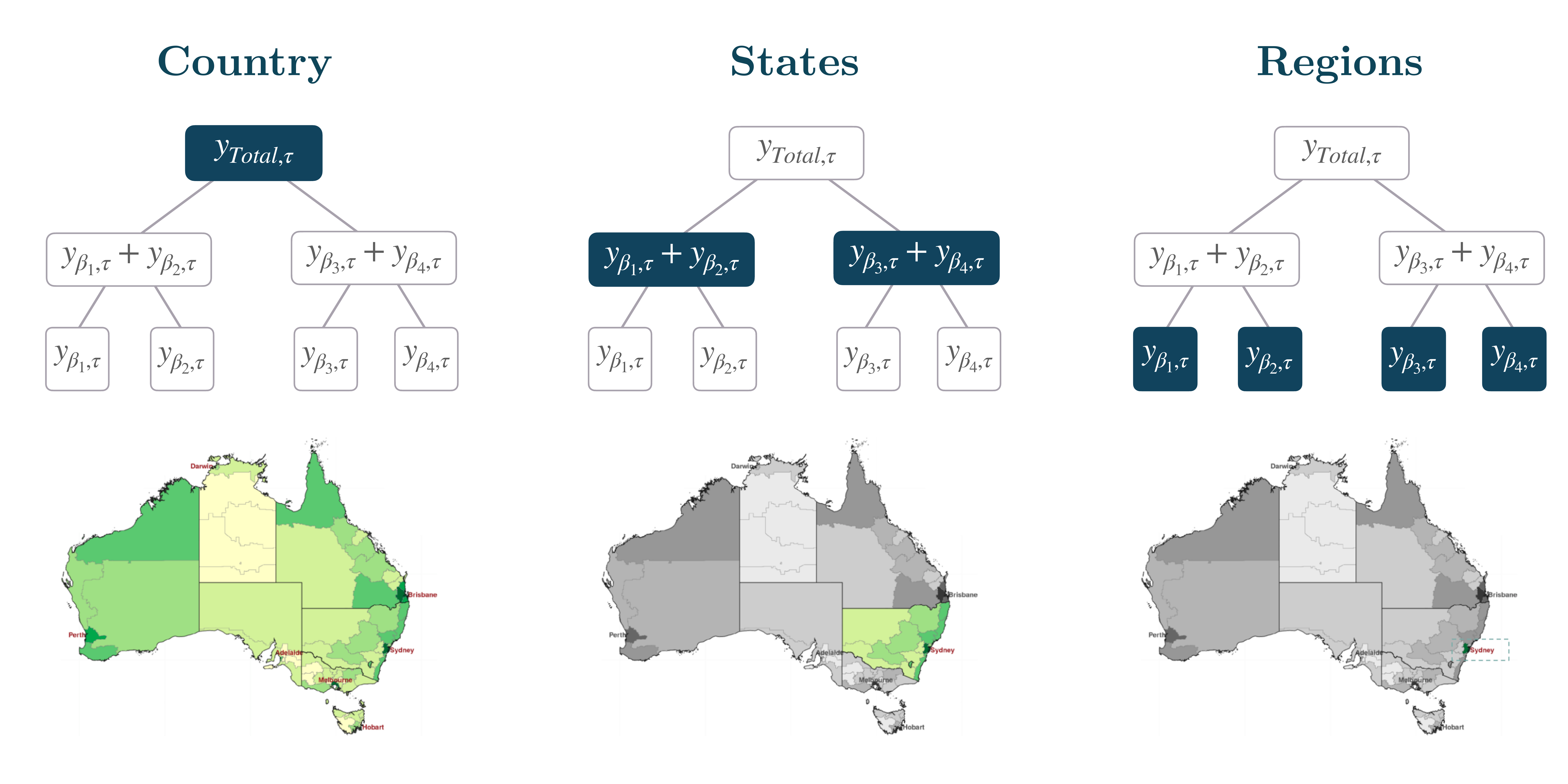}
\caption{A simple example of a four bottom-level series hierarchical structure with geographic levels: the top level corresponds to the country, the middle level to states, and the bottom level to regions.
}
\label{fig:hierarchical_example2}
\end{figure*}

\subsection{Hierarchical Reconciliation}

Classic hierarchical forecasting methods involve a two-stage process in which a set of univariate statistical base forecasts are reconciled. There is a long literature on these reconciliation strategies that include 
\BottomUp~\citep{orcutt1968hierarchical_bottom_up, dunn1976hierarchical_bottom_up2}, \TopDown~\citep{gross1990hierarchical_top_down, fliedner1999hierarchical_top_down2}, and more recent 
optimal reconciliation strategies like \Comb~\citep{hyndman2011optimal_combination_hierarchical}, \MinTrace~\citep{wickramasuriya2019hierarchical_mint_reconciliation}
and \ERM~\citep{taieb2019hierarchical_regularized_regression}. Probabilistic forecast reconciliation is at the forefront of hierarchical forecasting research. Among the few methods capable of probabilistic coherence, there is \PERMBU\ that infuses multivariate dependencies to bottom-level probabilities using copulas~\citep{taieb2017coherent_prob_forecasts}, \NORMALITY\ that reconciles a multivariate model under Gaussian assumptions~\citep{wickramasuriya2023probabilistic_gaussian}, and 
\BOOTSTRAP\ that generates reconciled forecasts with bootstrap reconciliation~\citep{panagiotelis2023probabilistic_reconciliation}.

\subsection{Hierarchical Neural Forecasting}

Neural network based methods have gained popularity in forecasting applications, outperforming most alternatives. As surveys show, in recent years, the academic community has greatly renovated interest in the topic \citep{benidis2020dl_timeseries_review2}. The literature has permeated into hierarchical forecasting, with contributions such as \SHARQ~\citep{han2021hierarchical_sharq}, \HIRED~\citep{paria2021hierarchical_hired}, and \PROFHIT~\citep{kamarthi2022profhit_network} approximate coherent methods using variants of bottom-up aggregation regularization. Fully coherent approaches include \HierEtoE~\citep{rangapuram2021hierarchical_e2e} a multivariate approach that incorporates \MinTrace-like reconciliation in the network's optimization, \TDProb~\citep{paria2022td_prob} that learns \TopDown\ proportions to probabilistically reconcile univariate base models.

Despite recent progress in extending neural networks toward hierarchical forecasting, existing solutions still face challenges: (i) their implementations rely on restrictive probabilistic assumptions or are not entirely coherent; (ii) computational complexity of multivariate approaches scales poorly; (iii) the forecast accuracy improvements over statistical baselines are still modest.

\subsection{Mathematical Notation}
A hierarchical time series (HTS) is a multivariate time series under aggregation constraints. We denote the HTS by the vector $\mathbf{y}_{[i],t} = [\,\mathbf{y}^{\intercal}_{[a],t}\,|\,\mathbf{y}^{\intercal}_{[b],t}\,]^{\intercal} \in \mathbb{R}^{N_{a}+N_{b}}$, for time step $t$, where $[a],[b]$ denote respectively the aggregate and bottom level indices. The total number of series in the hierarchy is $|[i]| = (N_{a}+N_{b})=N_{i}$. We distinguish between the time indices $[t]$ and forecast indices $\tau \in [t+1:t+h]$, and hierarchical, bottom and aggregate indexes $\iota \in [i], \beta \in [b]$ , $\alpha \in [a]$.

At any time $t$, the constraints are $\mathbf{y}_{[a],t} = \mathbf{A}_{[a][b]}  \mathbf{y}_{[b],t}$ where $\mathbf{A}_{[a][b]}$ denotes the relationship between the bottom-level series to the upper-level series. We can write the HTS as
\begin{equation}\label{eqn:summation}
\mathbf{y}_{[i],t}  = \mathbf{S}_{[i][b]} \mathbf{y}_{[b],t}
\quad \Leftrightarrow \quad 
\begin{bmatrix}\mathbf{y}_{[a],t}
\\
\mathbf{y}_{[b],t}\end{bmatrix} 
= \begin{bmatrix}
\mathbf{A}_{[a][b]}\\
\mathbf{I}_{[b][b]}
\end{bmatrix}\mathbf{y}_{[b],t}
\end{equation}
where $\mathbf{S}_{[i][b]}$ and $\mathbf{I}_{[b][b]}$ are summing and identity matrices. Figure~\ref{fig:hierarchical_example2} exemplifies $N_{i}=7$, $N_{b}=4$ and $N_{a}=3$:
\begin{equation}
\begin{split}
    \mathbf{y}_{[a],t} &=\left[y_{\mathrm{Total},t},\; y_{\beta_{1},t}+y_{\beta_{2},t},\;y_{\beta_{3},t}+y_{\beta_{4},t}\right]^{\intercal},  \\
    \mathbf{y}_{[b],t} &=\left[ y_{\beta_{1},t},\; y_{\beta_{2},t},\; y_{\beta_{3},t},\; y_{\beta_{4},t} \right]^{\intercal},
\end{split}
\end{equation}
where $y_{\mathrm{Total},t} = y_{\beta_{1},t}+y_{\beta_{2},t}+y_{\beta_{3},t}+y_{\beta_{4},t}$. The summing matrix associated to Figure~\ref{fig:hierarchical_example2} is:
\begin{equation}
\mathbf{S}_{[i][b]}
=
\begin{bmatrix}
           \\
\mathbf{A}_{\mathrm{[a][b]}} \\ 
           \\ 
           \\
\mathbf{I}_{\mathrm{[b][b]}} \\
           \\
\end{bmatrix}
=
\begin{bmatrix}
\;1 & 1 & 1 & 1 \\
\;1 & 1 & 0 & 0 \\
\;0 & 0 & 1 & 1 \\ \hline
\;1 & 0 & 0 & 0 \\
\;0 & 1 & 0 & 0 \\
\;0 & 0 & 1 & 0 \\
\;0 & 0 & 0 & 1 \\
\end{bmatrix}.
\end{equation}

\begin{definition}
\label{def:probabilistic_coherence} (Probabilistic Coherence).
Let $(\Omega_{[b]}, \mathcal{F}_{[b]}, \mathbb{P}_{[b]})$ be a probabilistic forecast space, with $\mathcal{F}_{[b]}$ a $\sigma$-algebra on $\mathbb{R}^{N_{b}}$. Let $\mathbf{S}(\cdot):\Omega_{[b]} \mapsto \Omega_{[i]}$ be the constraints' implied transformation. A coherent probabilistic forecast space $(\Omega_{[i]}, \mathcal{F}_{[i]}, \mathbb{P}_{[i]})$ satisfies:
\begin{equation}
    \mathbb{P}_{[i]}\left(\mathbf{S}(\mathcal{B})\right) = \mathbb{P}_{[b]}\left(\mathcal{B}\right) 
    \text{for set } \mathcal{B} \in \mathcal{F}_{[b]} \text{, image } \mathbf{S}(\mathcal{B}) \in \mathcal{F}_{[i]} 
\end{equation}

that is, it assigns a zero probability to any set without coherent forecasts \citep{panagiotelis2023probabilistic_reconciliation}.
\end{definition}

\begin{definition}
\label{def:hierarchical_reconciliation} (Hierarchical Reconciliation). 
For time $t$, horizon $h$, and forecast indexes $\tau \in [t+1:t+h]$. 
Reconciliation for point forecasts $\hat{\mathbf{y}}_{[i],\tau}$, is denoted by:
\begin{equation}
\label{eqn:mean_reconciliation}
\tilde{\mathbf{y}}_{[i],\tau} 
= \mathbf{S}_{[i][b]} \mathbf{P}_{[b][i]} \hat{\mathbf{y}}_{[i],\tau}
= \mathbf{SP}(\hat{\mathbf{y}}_{[i],\tau})
\end{equation}
where $\mathbf{P}_{[b][i]}$ is defined by the reconciliation technique. And $\mathbf{SP}(\cdot): \Omega_{[i]} \mapsto \Omega_{[b]} \mapsto \Omega_{[i]}$ is the reconciliation's transformation \citep{hyndman2021forecasting_book}.
\end{definition}

\section{\ourstext\ Methodology} \label{section3:methodology}
\begin{figure*}[t] 
\centering
\includegraphics[width=1.0\linewidth]{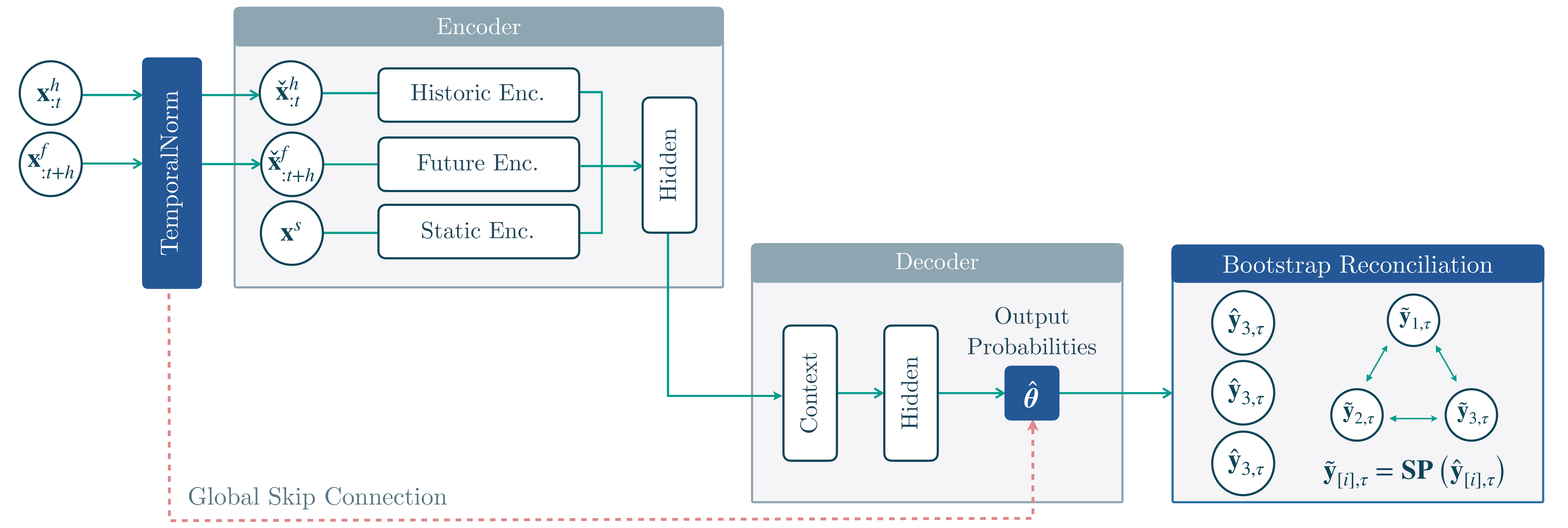}
\caption{
The \ours\ framework leverages a specialized multivariate mixture probability output that achieves coherency via bootstrap sample reconciliation $\mathbf{\tilde{y}}_{[i],\tau}=\mathbf{SP}(\mathbf{\hat{y}}_{[i],\tau})$. Additionally \ours\ incorporates the \TemporalNorm\ module to augment networks with scale-invariance through a global skip connection that enables input normalization and output scale recomposition.
}
\label{fig:hint_architecture}
\end{figure*}

The \ours\ framework estimates the following conditional probability under coherency constraints:
\begin{equation}
    \mathbb{P}\left(\mathbf{y}_{[i][t+1:t+h]}\,|\,\btheta \,\right)
\end{equation}

where $\btheta$ depends on  
$
\{
  \mathbf{x}^{h}_{[i][:t]},\;
  \mathbf{x}^{f}_{[i][:t+h]},\;
  \mathbf{x}^{s}_{[i]} \}$
historic, future and static variables. Here we describe our proposed approach, its main principles of operation are depicted in Figure~\ref{fig:hint_architecture}. A \ours\ network consists of a coherent probability output and a scale-robustified neural forecast architecture.

\subsection{Coherent Multivariate Mixture} \label{section:multivariate_mixture}

\ours\ is a highly modular system that supports a wide range of probabilistic outputs. We leverage \ours's flexibility to accommodate a multivariate Gaussian mixture model specialized in hierarchical forecasting. Its conditional forecast distribution is described by:
\begin{equation}
\begin{split}
\label{eqn:hierarchical_mixture}
    \hat{\mathbb{P}}\left(\mathbf{y}_{[i][t+1:t+h]}|\,\hat{\btheta}\right) \qquad \qquad \qquad \qquad \qquad \qquad\\ = 
    \sum_{\kappa=1}^{N_{k}} \hat{w}_\kappa \prod_{(\iota,\tau) \in [i][t+1:t+h]}
    \mathcal{N} \left(y_{\iota,\tau}|\;\hat{\mu}_{\iota,\kappa,\tau}\;\hat{\sigma}_{\iota,\kappa,\tau} \right)
\end{split}
\end{equation}

This multivariate mixture has advantageous theoretical properties, proven in Appendix~\ref{section:hierarchical_mixture_properties}. It can arbitrarily approximate univariate distributions and describe the series' correlations.

\textbf{Optimization:} \ours\ achieves high computational efficiency, because we optimize it through composite-likelihood~\citep{lindsay_1988, varin2011composite_likelihood} of the series in the SGD batches approximating the full joint distribution. Let $\mathcal{B}=\{[b_i]\}$ be time-series SGD batches, and $\bomega$ the neural network parameters, then \ours's negative log composite likelihood is:
\begin{equation}\label{eqn:composite_likelihood1}
\begin{split}
    \mathcal{L}(\bomega) = - \sum_{[b_{i}] \in \mathcal{B}}
    \mathrm{log} \Bigg[\sum_{\kappa=1}^{N_{k}} \hat{w}_\kappa(\bomega) \prod_{(\iota,\tau) \in [b_{i}][t+1:t+h]} \qquad \qquad \\
	    \left( 
	    \frac{1}{\hat{\sigma}_{\iota,\kappa,\tau}(\bomega) \sqrt{2 \pi}}
	    \exp{ \bigl\{
	    -\frac{1}{2}
	    \left(
	    \frac{y_{\iota,\tau}-\hat{\mu}_{\iota,\kappa,\tau}(\bomega)}{ \hat{\sigma}_{\iota,\kappa,\tau}(\bomega) } 
	    \right)^{2}
	    \bigl\} }
	    \right)     
    \Bigg]
\end{split}
\end{equation}

\textbf{Bootstrap Reconciliation:} We ensure the coherence via bootstrap reconciliation~\citep{panagiotelis2023probabilistic_reconciliation}.  Figure~\ref{fig:motivation}.c shows it restoring the aggregation constraints into the base samples. Let $\mathcal{H}$ be a coherent forecast set, and $\mathbf{SP}^{-1}(\cdot)$ a reconciliation's inverse image, Theorem~\ref{theorem:analytical_reconciled_probability} analytically derives the coherent forecast distribution:
\begin{equation}
    \tilde{\mathbb{P}}\left(\tilde{\mathbf{y}}_{[i],\tau}\in\mathcal{H}|\,\tilde{\btheta}\right)
    =
    \hat{\mathbb{P}}\left(\hat{\mathbf{y}}_{[i],\tau} \in \mathbf{SP}^{-1}(\mathcal{H})|\,\hat{\btheta}\right)
\end{equation}

Prior research has used properties of base forecast distributions and analytically reconciled probabilities to achieve efficient inference times~\citep{olivares2022hierarchicalforecast}. 

The \emph{bootstrap reconciliation} technique offers a distribution-agnostic reconciliation method. We leverage this flexibility to study \ours\ with different distributions outputs in our ablation studies in Figure~\ref{fig:distribution_ablation}.

\begin{theorem} \label{theorem:analytical_reconciled_probability}
Consider a reconciliation from Definition~\ref{def:hierarchical_reconciliation} where the entire hierarchy's forecasts are combined into reconciled bottom-level forecasts using a composition of linear transformations $\mathbf{SP}(\cdot)=\mathbf{S_{[i][b]}P_{[b][i]}}(\cdot)$. The reconciled probability for the entire hierarchical series is given by:
\begin{equation}\label{eqn:reconciled_hierarchical1}
\begin{split}
\tilde{\mathbb{P}}_{[b]}\left(
\mathbf{y}_{[b],\tau}\right) = 
|\mathbf{P}^{*}| \int \hat{\mathbb{P}}_{[i]}\left(\mathbf{P}_{\perp} \mathbf{y}_{[a],\tau} +  \mathbf{P}^{-} \mathbf{y}_{[b],\tau}\right) \delta \mathbf{\mathbf{y}_{[a],\tau}} \\
\tilde{\mathbb{P}}
\left(
\mathbf{y}_{[i],\tau}\right) =|\mathbf{S}^{*}| \tilde{\mathbb{P}}_{[b]}\left(\mathbf{S}^{-} \mathbf{y}_{[i],\tau} \right) \mathbbm{1}\{\mathbf{y}_{[i],\tau} \in \mathcal{H}\} \qquad \qquad
\end{split}    
\end{equation}

where $\tilde{\mathbb{P}}_{[b]}(\cdot)$ is the reconciled bottom forecast distribution, $\mathbbm{1}(\tilde{\mathbf{y}}_{[i],\tau} \in \mathcal{H})$ indicates if realization belongs in the $N_{b}$-dimensional hierarchically coherent subspace $\mathcal{H}$, $\mathbf{P}^{-}\in \mathbb{R}^{N_{i} \times N_{b}},\mathbf{S}^{-} \in \mathbb{R}^{N_{b} \times N_{i}}$ are $\mathbf{P}_{[b][i]},\mathbf{S}_{[i][b]}$ Moore-Penrose inverses and $\mathbf{P}_{\perp} \in \mathbb{R}^{N_{a} \times N_{i}},\mathbf{S}_{\perp} \in \mathbb{R}^{N_{i} \times N_{a}}$ their orthogonal complements, $\mathbf{P}^{*} = \left[\mathbf{P}_{\perp} \,|\, \mathbf{P}^{-} \right]$, and $\mathbf{S}^{*}=\left[\mathbf{S}^{-}_{\perp} \,|\, \mathbf{S} \right]^{-1}$. Proof available in Appendix~\ref{section:hierarchical_mixture_properties}.
\end{theorem}

\begin{theorem} \label{theorem:multivariate_mixture_covariance}
The multivariate mixture from Equation~(\ref{eqn:hierarchical_mixture}) models the time series' relationships through the structure of its latent variables, its non-diagonal covariance is the following $N_{k}-1$ rank matrix:
\begin{equation}
\label{eqn:covariance_multivariate1}
    \mathrm{Cov}(\mathbf{Y}_{[i],\tau}) = 
    \sum^{N_{k}}_{\kappa=1} \hat{\mathbf{w}}_{\kappa}
    (\hat{\bmu}_{[i],\kappa,\tau}-\bar{\bmu}_{[i],\tau})
    (\hat{\bmu}_{[i],\kappa,\tau}-\bar{\bmu}_{[i],\tau})^{\intercal} 
\end{equation}
Proof in Appendix~\ref{section:hierarchical_mixture_properties}.
\end{theorem}

Figure~\ref{fig:forecast_comparison} shows the importance of the flexibility of the multivariate mixture. When combined with the \BottomUp\ reconciliation strategy, a univariate independence simplification of Equation~(\ref{eqn:hierarchical_mixture}) translates into overestimating the aggregate forecast's uncertainty. 

Estimating the series' relationships can help sharpen the reconciled forecast distribution. Unlike previous research on high-dimensional multivariate techniques with covariance matrix low-rank approximations~\citep{salinas2019low_rank_covariance}, our distribution does not store the covariance for greatly improved memory complexity.
\begin{equation}
    \mathcal{O}(\mathrm{batch\;size}) \text{ vs } \mathcal{O}(N^{2}_{i})
\end{equation}

\begin{figure}[!ht] 
\centering     
\includegraphics[height=3.2 cm]{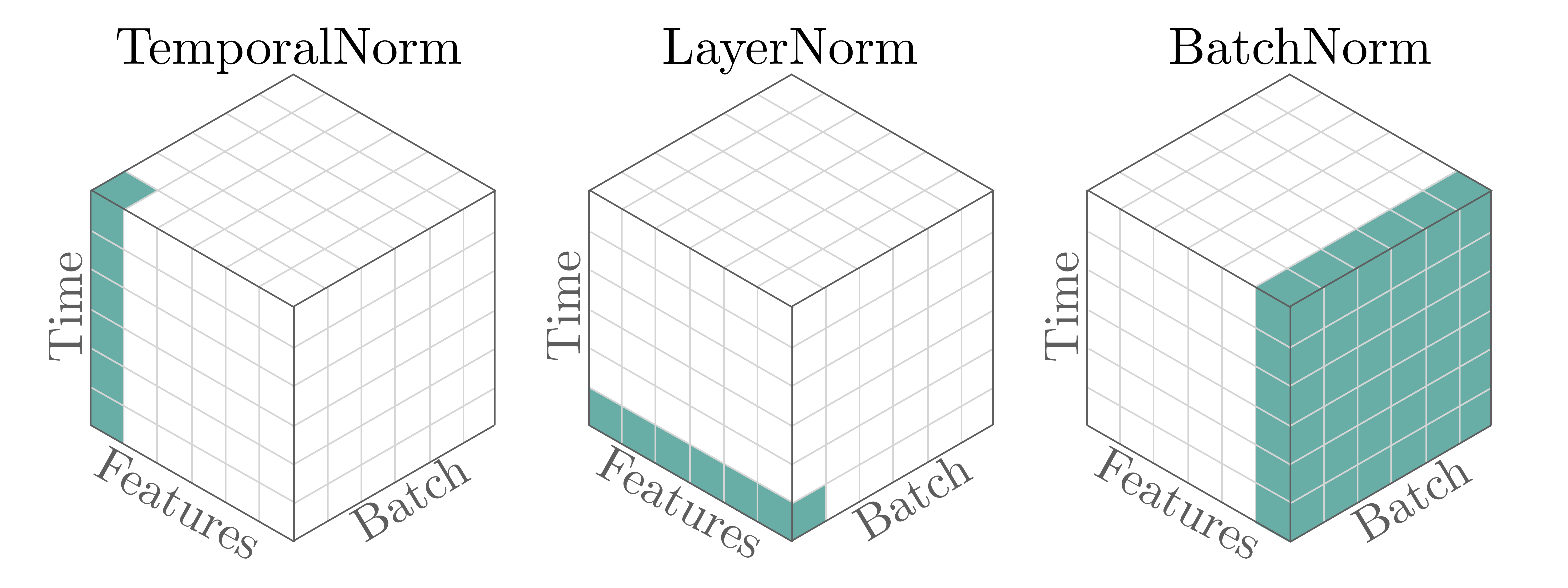}
\caption{Temporal normalization (left), layer normalization (center) and batch normalization (right). The entries in green are in the normalizing statistics. }
\label{fig:temporal_norm}
\end{figure}

\subsection{Probabilistic Reversible Instance Normalization} \label{section:scale_decouple}

Well-performing neural architectures, such as \TFT~\citep{lim2021tft}, \DeepAR~\citep{salinas2020deepAR}, and \NBEATSx~\citep{oreshkin2020nbeats,olivares2022nbeatsx}, covertly incorporated scale-robustified cross-learning optimization. Our \TemporalNorm\ module standardizes the approach and makes it available to other architectures. \cite{kim2022reversible_normalization} recently proposed a general temporal normalization technique. We expanded on the reversible instance normalization approach by increasing the available scaling methods and extending them to probabilistic outputs.


\begin{figure*}[t]
\centering
\subfigure[\ours\ Univariate Mixture]{\label{fig:a}
\includegraphics[width=66mm]{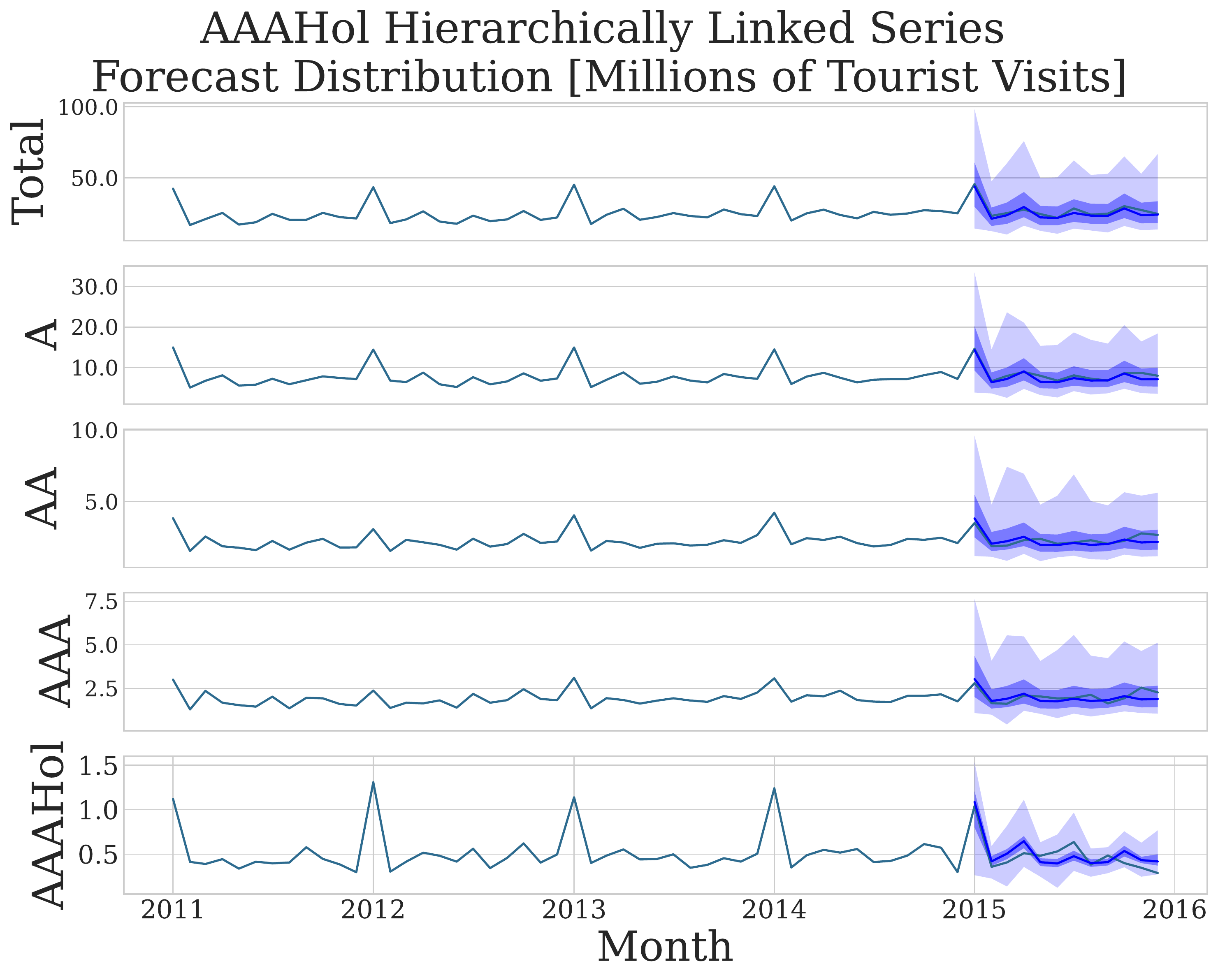}
\label{fig:univariate_reconciliation}
}
\subfigure[\ours\ Multivariate Mixture]{\label{fig:b}
\includegraphics[width=66mm]{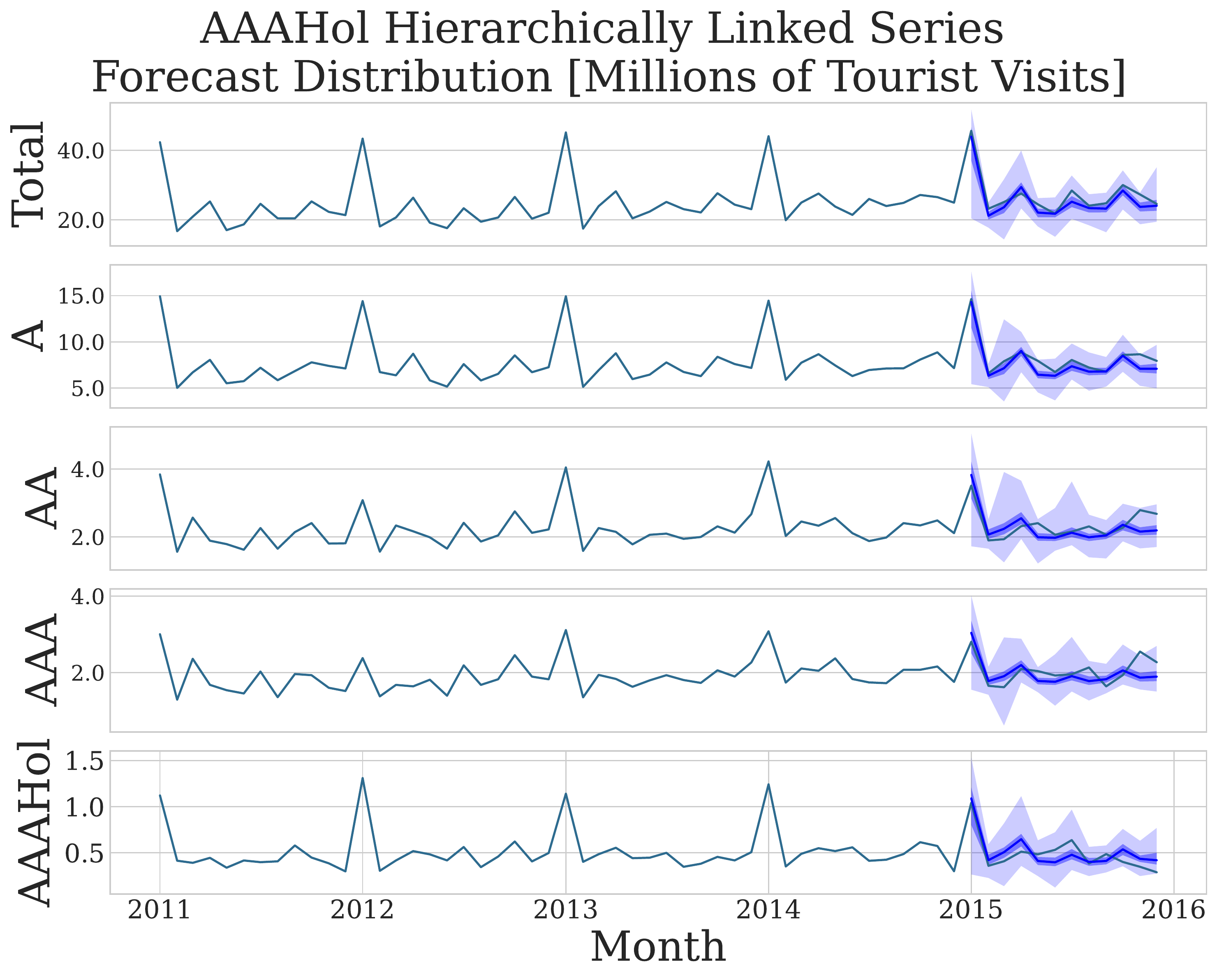}
\label{fig:multivariate_reconciliation}
}
\caption{
\ours\ univariate/multivariate estimation comparison. Different rows of plots show forecasts for different levels of  hierarchical aggregation. Light and dark blue represent 99\% and 75\% prediction intervals. Modeling the series' correlations sharpens the reconciled forecast distributions.}
\label{fig:forecast_comparison}
\end{figure*}


\begin{table*}[t] 
\tiny
\centering
\caption{
Empirical evaluation of probabilistically coherent forecasts. Scaled continuous ranked probability score (sCRPS) and relative mean squared error (relMSE), averaged over 10 random seeds, over all hierarchy series. The best results are bold-faced (lower values of metrics used are preferred).
\tiny{
\textsuperscript{\dag} The \PROFHIT\ results differ from Kamarthi et al. 2022, as the only available implementation appears to be numerically unstable in its optimization. $\ddagger$ \PatchTST/\NBEATS\ use deterministic coherent normality intervals. \textsuperscript{*} Best performing variant of \TopDown\ (avg.\ proportions, proportions avg.), and \MinTrace\ (ols, wls, shrinkage) reported. \textsuperscript{**} \PERMBU\ only available for strict hierarchies.
}
}
\label{table:summarized_crps_evaluation}
\setlength\tabcolsep{2.3 pt}
\begin{tabular}{cl|cc|cccc|cc|cc}
\toprule
                       & &   \multicolumn{2}{c|}{\ours~(Ours)}     &     \multicolumn{4}{c|}{Neural Forecast}                             &      \multicolumn{2}{c|}{\ARIMA-\BOOTSTRAP}                                          &  \multicolumn{2}{c}{\ARIMA-\PERMBU\textsuperscript{**}} \\
    & \textsc{Dataset}   & Coherent &      Base                  &\HierEtoE   &    \PROFHIT\textsuperscript{\dag} & \PatchTST$\ddagger$   & \NBEATS$\ddagger$       &    \BottomUp &  \MinTrace\textsuperscript{*}  &       \BottomUp  &          \MinTrace\textsuperscript{*}  \\ \midrule
\parbox[t]{.4mm}{\multirow{5}{*}{\rotatebox[origin=c]{90}{sCRPS}}}
    & \Labour            & \textbf{.0067±.0001}  &   .0070±.0001  &  .0171±.0003          &    .2138±.0070        & \EDIT{0.0094}      & \EDIT{0.0077}         &  .0078±.0001  &  .0073±.0000                   &  .0077±.0001  &  .0069±.0001       \\
    & \Traffic           & .0589±.0004           &   .0633±.0003  &  \textbf{.0426±.0008} &    .1137±.0022        & \EDIT{0.0872}      & \EDIT{0.0781}         &  .0736±.0024  &  .0608±.0014                   &  .0849±.0009  &  .0651±.0008       \\
    & \Tourism           & \textbf{.0536±.0004}  &   .0574±.0004  &  .0761±.0007          &    .1358±.0033        & \EDIT{0.1093}      & \EDIT{0.0736}         &  .0682±.0018  &  .0703±.0017                   &  .0649±.0016  &  .0680±.0016       \\
    & \TourismL          & \textbf{.1176±.0002}  &   .1253±.0002  &  .1424±.0019          &    .2139±.0014        & \EDIT{0.1277}      & \EDIT{0.1291}         &  .1375±.0013  &  .1313±.0009                   &  -            &  -                 \\
    & \Wikitwo           & \textbf{.2447±.0007}  &   .2395±.0006  &  .2592±.0031          &    .4009±.0028        & \EDIT{0.2999}      & \EDIT{0.3621}         &  .2894±.0038  &  .2808±.0035                   &  .3920±.0044  &  .3821±.0049       \\ \midrule
\parbox[t]{.4mm}{\multirow{5}{*}{\rotatebox[origin=c]{90}{relMSE}}}
    & \Labour            & .5802±.0131           &   .6333±.0141  & .8165±0.0353          & $6.774 \times 10^{3}$ & \EDIT{0.6132}      & \EDIT{0.4932}         & 0.5382±.0000  & \textbf{0.3547±.0000}          &               &                    \\
    & \Traffic           & .1212±.0051           &   .1340±.0062  & \textbf{.0328±0.0019} & 0.4536±.0224          & \EDIT{0.1589}      & \EDIT{0.1489}         & 0.1392±.0000  & 0.0744±.0000                   &               &                    \\
    & \Tourism           & \textbf{.0387±.0007}  &   .0574±.0008  & .1471±0.0046          & 0.9745±.0803          & \EDIT{0.0671}      & \EDIT{0.0585}         & 0.1002±.0000  & 0.1235±.0000                   &               &                    \\
    & \TourismL          & \textbf{.0577±.0009}  &   .0660±.0010  & .2449±0.0096          & 1.0401±.0296          & \EDIT{0.0742}      & \EDIT{0.0692}         & 0.3070±.0000  & 0.1375±.0000                   &               &                    \\
    & \Wikitwo           & \textbf{.1884±.0012}  &   .1966±0.0017 & .6598±0.0249          & 0.7901±.0384          & \EDIT{0.2014}      & \EDIT{0.2137}         & 1.0163±.0000  & 1.0068±.0000                   &               &                    \\ \bottomrule
\end{tabular}
\end{table*}

\section{Experimental Results} \label{section4:evaluation}


\textbf{Hierarchical Forecast Datasets.} We follow experimental protocols established in previous research by \citet{rangapuram2021hierarchical_e2e}. The benchmark datasets are: Monthly Australian \Labour~\citep{Aulabor2019Aulabor_dataset}, SF Bay Area daily \Traffic~\citep{dua2017traffic_dataset},
Quarterly Australian \Tourism\ visits~\citep{canberra2005tourismS_dataset}, Monthly Australian \TourismL\ visits~\citep{canberra2019tourismL_dataset}, and daily \Wikitwo\ views~\citep{wikipedia2018web_traffic_dataset}. Appendix~\ref{section:hierarchical_datasets} includes a detailed exploration of this data.

\textbf{Baselines.} In our main experiment, we compare with SoTA probabilistic coherent methods. Neural forecasting baselines include
(1) \HierEtoE~\citep{rangapuram2021hierarchical_e2e}, (2) \PROFHIT~\citep{kamarthi2022profhit_network}, 
(3) \PatchTST~\citep{nie2023patchTST}, (4) \NBEATSx~\citep{oreshkin2020nbeats,olivares2022nbeatsx},
while statistical baselines include variants of (5) \BOOTSTRAP~\citep{panagiotelis2023probabilistic_reconciliation}, and (6) \PERMBU~ probabilistic reconciliation~\citep{taieb2017coherent_prob_forecasts} in combination with \BottomUp~\citep{orcutt1968hierarchical_bottom_up}, and \MinTrace~\citep{wickramasuriya2019hierarchical_mint_reconciliation} reconcilers. We use \HierarchicalForecast\ library baselines implementation~\citep{olivares2022hierarchicalforecast}. 

\textbf{Evaluation Metrics.} To assess the forecast accuracy of our method, we compute the scaled Continuous Ranked Probability Score (sCRPS; \citealt{matheson1976evaluation_crps, makridakis2022m5_uncertainty}) and the Relative Mean Squared Error (relMSE; \citealt{hyndman2006another_look_measures,olivares2021pmm_cnn}). Note that probabilistic coherence naturally implies mean hierarchical coherence.
\begin{equation}
\begin{split}
\mathrm{sCRPS}(\mathbb{P}, \mathbf{y}_{[i],\tau}) = \frac{2}{|[i\,]|} \sum_{i}
    \frac{\int^{1}_{0} \mathrm{QL}(\mathbb{P}_{i,\tau}, y_{i,\tau})_{q} dq }{\sum_{i} | y_{i,\tau} |} \qquad \\
    \mathrm{relMSE}(\mathbf{y}_{[i]}, \hat{\mathbf{y}}_{[i]}, \mathbf{\check{y}}_{[i]}) =
    \frac{\mathrm{MSE}(\mathbf{y}_{[i]}, \mathbf{\hat{y}}_{[i]})}{\mathrm{MSE}(\mathbf{y}_{[i]}, \mathbf{\check{y}}_{[i]})}    \qquad\qquad\qquad 
\end{split}
\end{equation}
where $\mathrm{QL}(\hat{\mathbb{P}}_{i,\tau}, y_{i,\tau})_{q}$ stands for the q-quantile loss.



\begin{figure*}[t] 
    \centering
    \subfigure[\emph{Example of a Coherent Multivariate Mixture}]{
    \label{fig:mixture_components}
    \includegraphics[width=0.48\linewidth]
    {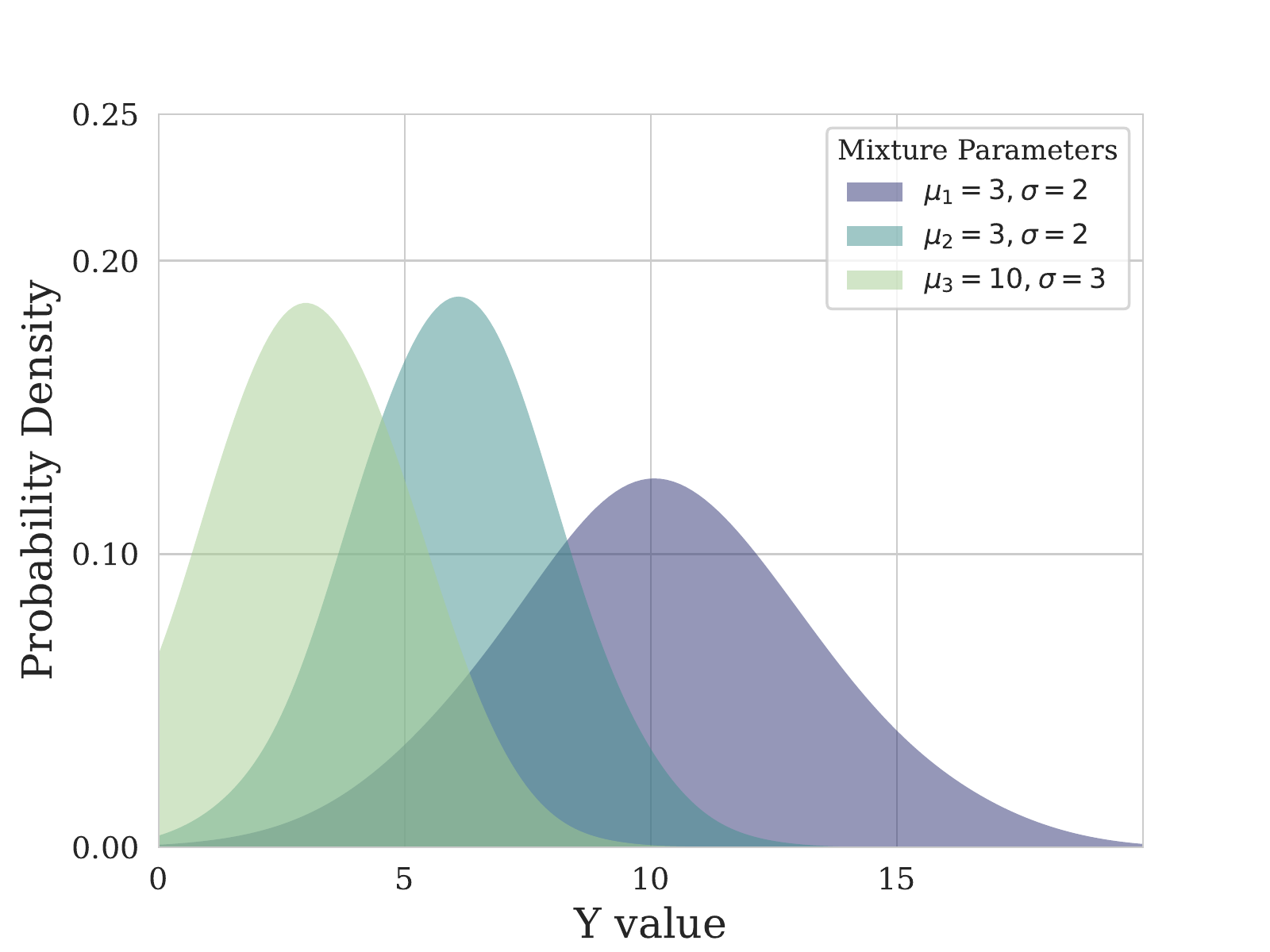}
    }
    \subfigure[\emph{Contributions of the Mixture Components}]{
    \label{fig:temporal_norm1}
    \includegraphics[width=0.44\linewidth]{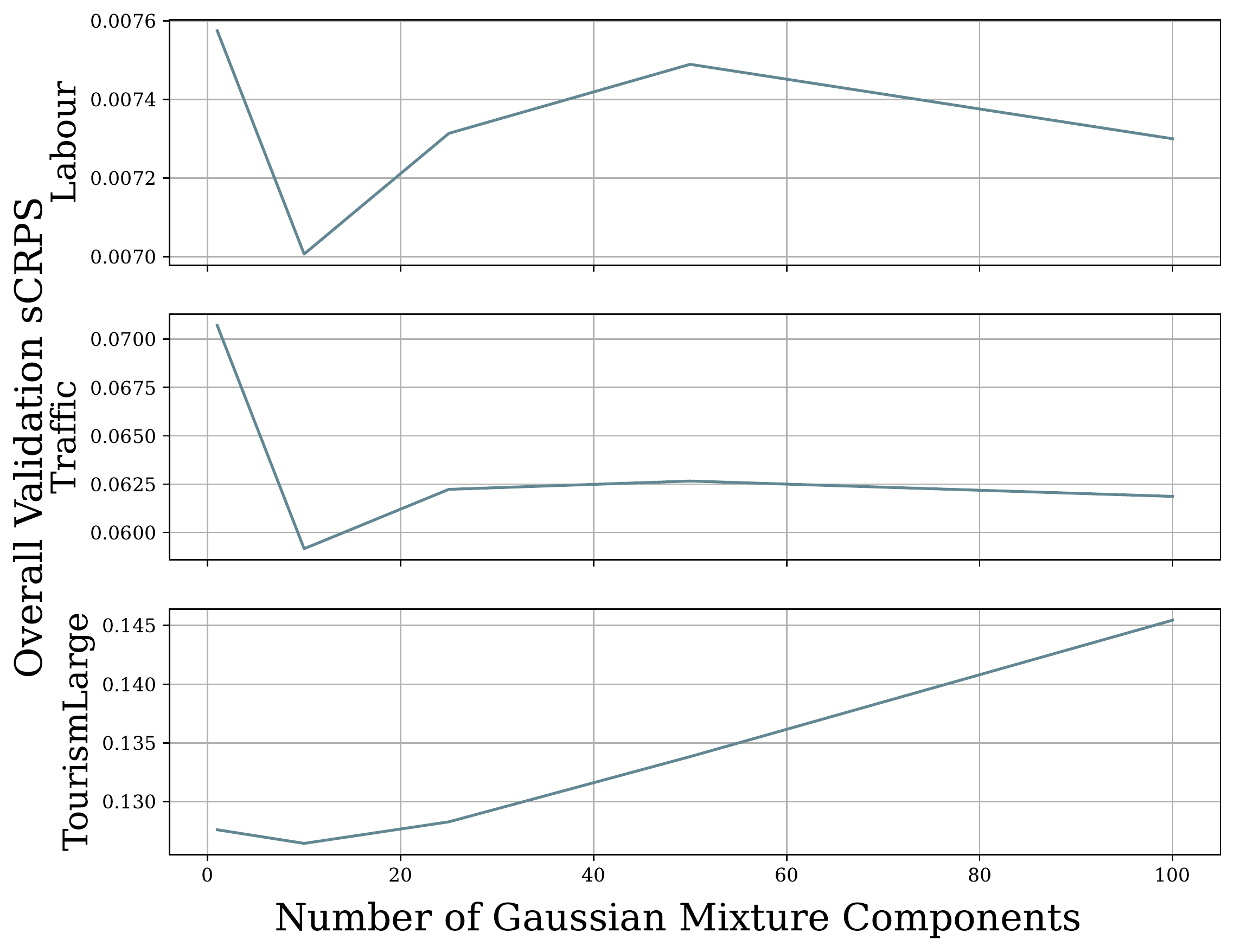} 
    }
    \caption{(a) Modularity of \ours\ can accommodate various distribution outputs and allows us to leverage hierarchically coherent multivariate mixture output module. (b) Validation sCRPS results as a function of mixture components. The sCRPS performance curves of \ours\ exhibit a bias-variance tradeoff shape, with an optimal number of components (10).}
    \label{fig:ablation}
\end{figure*}

\textbf{\ours\ configurations.}
We showcase \ours's modularity by augmenting three well-established neural forecast architectures: \NHITS~\citep{oreshkin2020nbeats,challu2022nhits}, \TFT~\citep{lim2021tft}, and \TCN~\citep{zico2018tcnn}. We optimally select \ours's underlying neural architecture using temporal cross-validation; Appendix~\ref{section:training_methodology} describes analyzed configurations and hyperparameter selection process.  
The experiments involve two alternative scenarios: the first is a base \ours\ with no reconciliation with no guaranteed coherent forecasts, the second is \ours\ with end-to-end reconciliation.

\subsection{Empirical Results}

\textbf{Main Results:} 
In Table~\ref{table:summarized_crps_evaluation}, \ours\ demonstrates superior performance across all datasets, except for \Traffic\, where it ranks as the second-best, behind only \HierEtoE. We execute all methods eight times and present its average and standard deviation of the sCRPS and relMSE scores. Overall \ours\ improves sCRPS by 13.2\% on average on all datasets except \Traffic. The relMSE results are highly correlated, although relMSE is more susceptible to outliers. We extend these results in Appendix~\ref{section:extended_main_results} where we explore the performance of the methods across levels of hierarchical aggregation.

\textbf{Ablation Studies:} We conducted ablation studies on different variants of our model. Appendix~\ref{section:scaled_decoupled_ablation}, examines the impact of the \TemporalNorm\ module. We observe that scale-decoupled optimization can improve accuracy of neural forecasting by an order of magnitude. Figure~\ref{fig:revin_ablation} also shows how robust scaling significantly improves over vanilla \REVIN~\citep{kim2022reversible_normalization}.

\begin{figure}[!ht] 
\centering     
\includegraphics[height=4.0 cm]{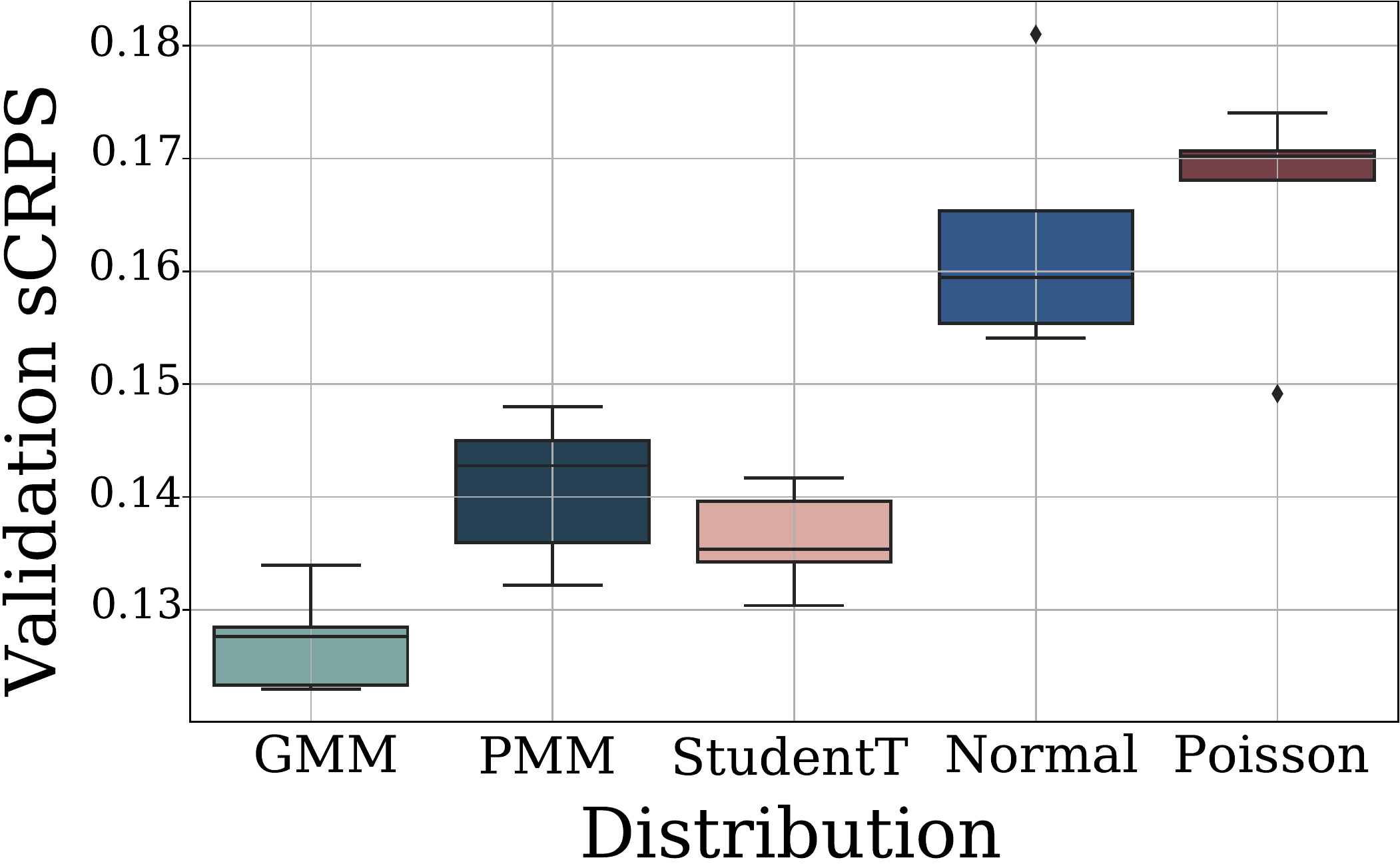}
\caption{Ablation study of forecast distribution's accuracy in the \TourismL\ dataset. Our multivariate mixture model translates into clear accuracy gains.}
\label{fig:distribution_ablation}
\end{figure}

Appendix~\ref{section:mixture_ablation} ablation study guides our selection for the number of mixture components, as shown in Figure~\ref{fig:ablation}, there is a clear advantage in the usage of a flexible multivariate mixture, as our method is both capable of improving over restrictive distribution assumptions and improve the forecast accuracy by capturing the relationships among series in the hierarchy. 

\begin{figure}[!ht] 
\centering     
\includegraphics[height=3.9 cm]{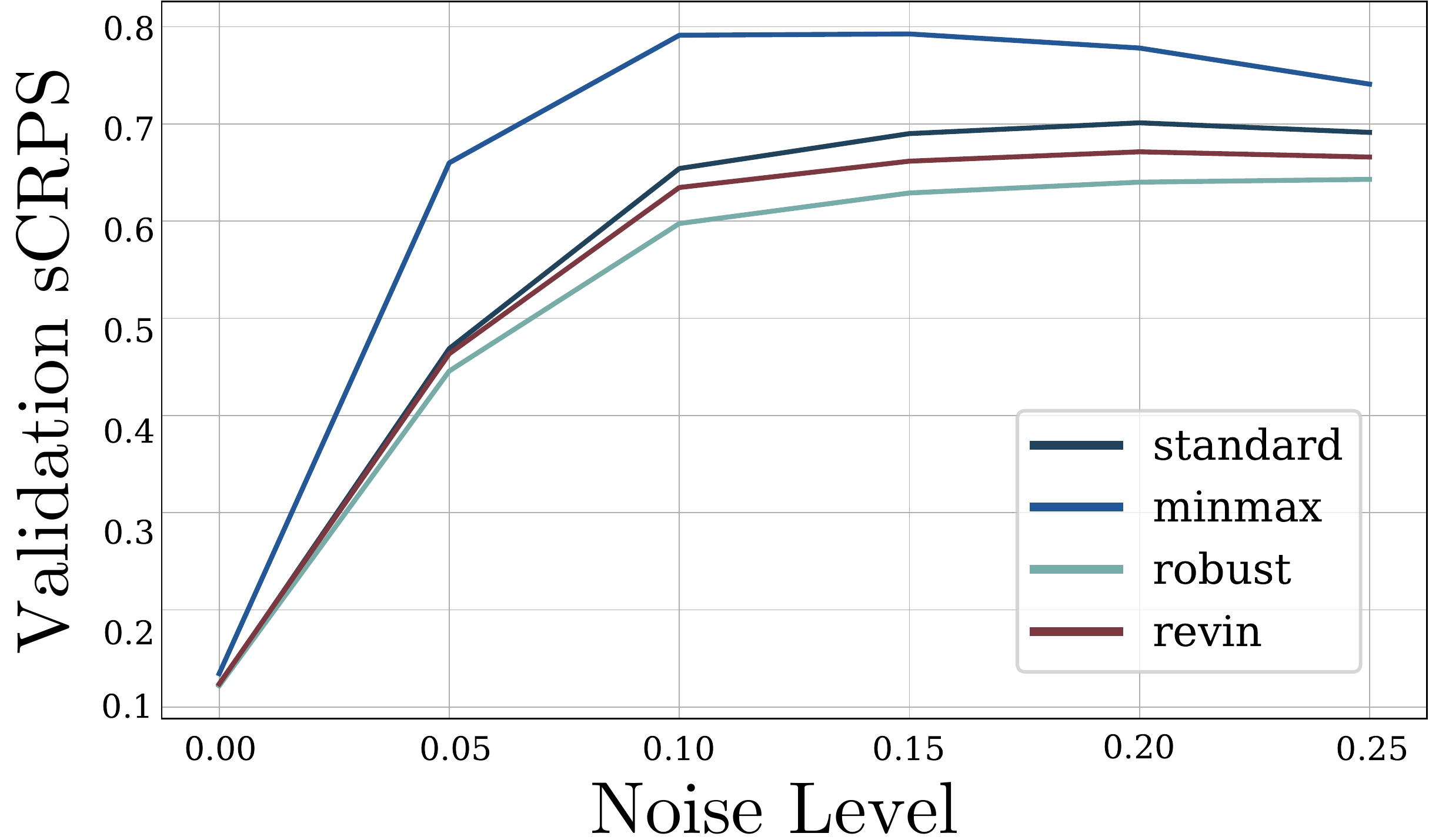}
\caption{Ablation study of scaling variants in the \TemporalNorm\ module and its effects on forecasts' accuracy in \TourismL\ dataset. In the presence of noise, robust scaling improves over \REVIN\ baseline.}
\label{fig:revin_ablation}
\end{figure}

Our ablation studies place the origins of the observed improvements in the \ours\ framework: (i) the mixture distribution improves performance upon simpler and more restrictive probabilistic output assumptions. (ii) scale-robustified optimization is a crucial enabler of cross-learning under stark scale variations; (iii) advanced reconciliation techniques can overcome potential systemic biases of base distributions; (iv) the latest neural forecast architectural innovations tend to improve the forecast accuracy.

\section{Future Research and Discussion} \label{section5:discussion}
The \ourstext\  (\ours) framework aims to learn the joint distribution for the entire hierarchy, while guaranteeing its coherence via bootstrap reconciliation.
The proposed framework is highly modular and enables many possible extensions, some of which we briefly discuss below.

Concerning hierarchical reconciliation, while most hierarchical forecast methods rely on post hoc techniques, the potential to learn the reconciliation method using a machine learning framework has received relatively little attention. However, initial findings along this line of investigation show very promising results \citep{taieb2019hierarchical_regularized_regression, das2023hierarchical_dirichlet}.

In our experiments, the only case where \ours\ did not outperform the \HierEtoE~\citep{rangapuram2021hierarchical_e2e} baseline was the \Traffic\ dataset. \HierEtoE\ outperforms all the other methods on those data due to including all the series as predictor variables in a VAR-like multivariate forecasting system under clear Granger causal relationships. Our results using univariate inputs challenge the effectiveness of existing multivariate input hierarchical forecasting approaches; the design of algorithms capable of maintaining good performance in the absence of Granger causalities while leveraging them, if present, is an exciting line of future research.

The demand for large-scale hierarchical forecasting systems is increasing across multiple important application domains. However, existing methods typically rely on access to the entire hierarchical structure for reconciliation. Innovative variations of the TopDown and BottomUp reconciliation approaches could be employed to develop a parallelizable approach that could leverage distributed computing frameworks.

\newpage
\section{Conclusion} \label{section6:conclusion}


We introduced \ours, a new framework that augments proven neural forecasting architectures to generate probabilistically coherent hierarchical forecasts. Two highly modular novelties are foundations of our approach: (i) a highly efficient multivariate mixture optimized with composite likelihood and transformed via bootstrap sample reconciliation, and (ii) incorporation of robust feature extraction and recomposition of probabilistic output scales within the network's architecture.

We empirically demonstrated that the flexibility of our multivariate mixture output can yield significant accuracy improvements (13.2\% on average on representative collection of benchmark data) compared to state-of-the-art baselines. These findings highlight the effectiveness of the presented approach in producing accurate and coherent forecasts and pave the way for further advancements in hierarchical forecasting. 

\ours\ is open-sourced, and practitioners can easily incorporate this framework into their work by accessing the code under this link~\href{https://anonymous.4open.science/r/HINT-559C/README.md}{\textcolor{blue}{http URL}}.

\clearpage
\bibliography{citations}
\bibliographystyle{plainnat}

\newpage
\onecolumn
\appendix
\setcounter{table}{0}
\setcounter{figure}{0}
\renewcommand{\thetable}{A\arabic{table}}

\section{Coherent Multivariate Mixture Properties}
\label{section:hierarchical_mixture_properties}

\subsection{Base Multivariate Mixture Probability}

\begin{figure*}[ht]
\label{fig:mixture_histograms}
\centering
\begin{tabular}{ccc}
    \subfigure[Gaussian Mixture]{\includegraphics[width=0.47\textwidth]{images/fig4a_gmm_sample.pdf}} 
    &
    \subfigure[Poisson Mixture]{\includegraphics[width=0.47\textwidth]{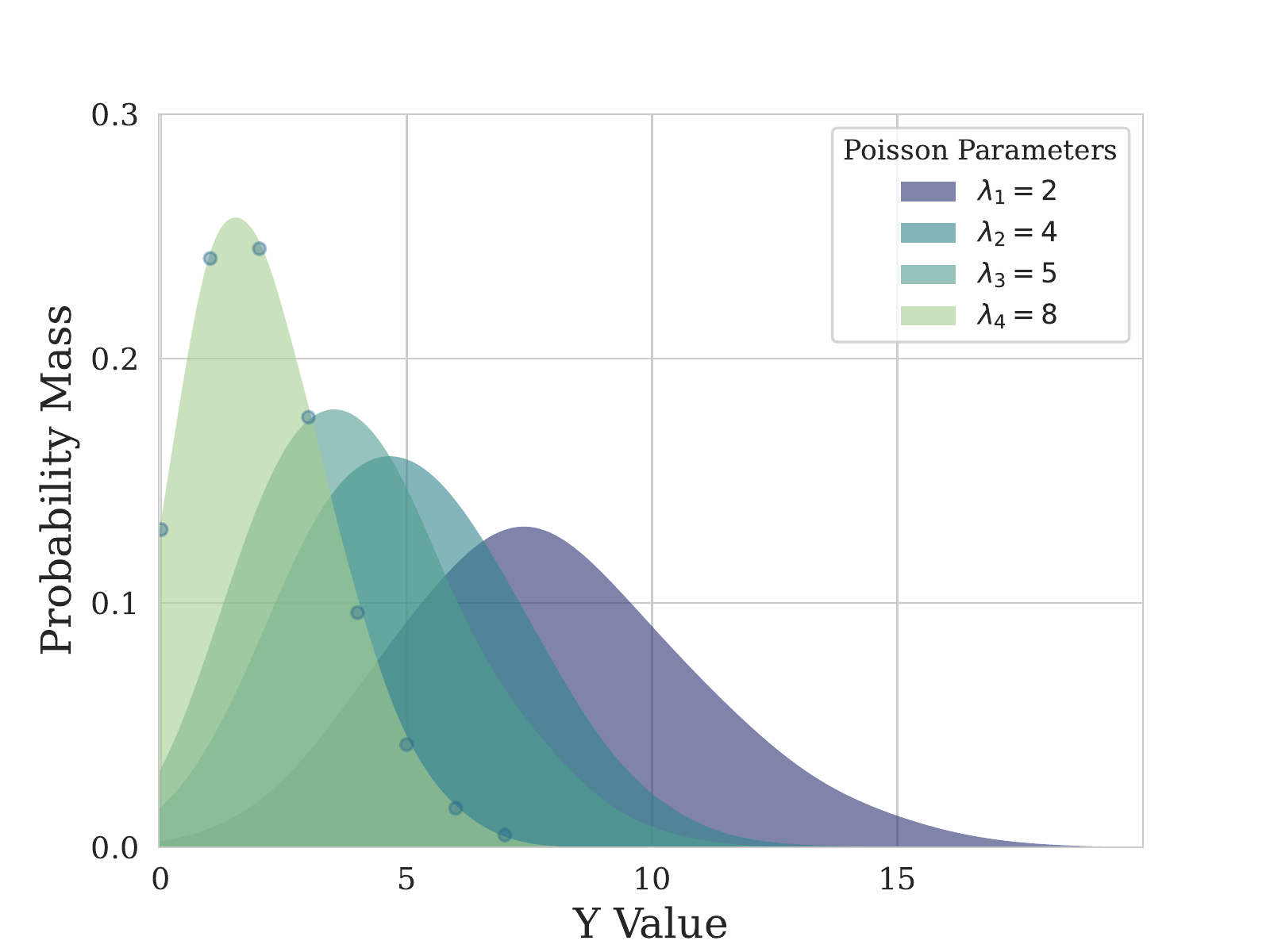}} 
\end{tabular}
\caption{\ours's multivariate joint distribution has advantageous properties that make it uniquely suited for hierarchical forecasting.
It is highly flexible, capable of efficiently modeling series' relationships, and under minimal restrictions, guarantees probabilistic coherence.}
\end{figure*}

In Section~\ref{section3:methodology}, we highlight that \ours\ boasts a flexible and modular framework that can handle various probabilistic outputs. We use this flexibility to enhance the networks with a multivariate mixture density. Specifically, the mixture model describes the joint probability of the hierarchical time series $\mathbf{Y}_{[i][t+1:t+h]}$ as follows:
\begin{equation*}
\label{eqn:hierarchical_mixture2}
    \mathbb{P}\left(Y_{[i][t+1:t+h]}=\mathbf{y}_{[i][t+1:t+h]}|\,\hat{\btheta}\right)
    = \sum_{\kappa=1}^{N_{k}} \hat{w}_\kappa \prod_{(\iota,\tau) \in [i][t+1:t+h]} 
    \mathcal{N} \left(y_{\iota,\tau}|\;\hat{\mu}_{\iota,\kappa,\tau}\;\hat{\sigma}_{\iota,\kappa,\tau} \right)
\end{equation*}

where the mixture describes individual series through the location and variance parameters.
For simplicity we denote the combined parameters $\hat{\btheta}_{[i][k][t+1:t+h]} = [\hat{\bmu}_{[i][k][t+1:t+h]}\;|\;\hat{\bsigma}_{[i][k][t+1:t+h]}]$.

Under reasonable assumptions for the underlying probability, the mixture distribution offers arbitrary approximation guarantees~\citep{titterington1985statistical,nguyen2018mixture_approximations}. We can control its flexibility by adjusting the number of components $|[k]|=N_{k}$. Furthermore, the mixture is not limited to Gaussian components; we can extend it to include discrete variables. Figure~\ref{fig:mixture_histograms} presents an example of its marginal probabilities.

\textbf{Conditional Independence:} A key consequence of the multivariate mixture probability in Equation~(\ref{eqn:hierarchical_mixture2}), is the assumption that the modeled series $\mathbf{y}_{[i][t+1:t+h]}$ are conditionally independent given the parameters $\hat{\bmu}_{[i][k][t+1:t+h]}$ and $\hat{\bsigma}_{[i][k][t+1:t+h]}$.  That is for any series and horizons $ (\iota, \tau) \neq (\iota',\tau') \text{, } (\iota, \tau), (\iota', \tau') \in [b][t+1:t+h]$ and $\kappa \in [k]$:
\begin{equation}
\label{eqn:conditional_indep}
    \mathbb{P}(Y_{\iota,\tau},  Y_{\iota',\tau'} \vert \hat{\theta}_{\iota,\kappa,\tau},  \hat{\theta}_{\iota',\kappa,\tau'} )
    = \mathbb{P}(Y_{\iota,\tau} \vert \hat{\theta}_{\iota,\kappa,\tau}) \hat{\mathbb{P}}(Y_{\iota',\tau'},  \vert \hat{\theta}_{\iota',\kappa,\tau'}) 
\end{equation}

\textbf{Computational Efficiency:} To handle large-scale data scenarios, we explicitly avoid using a multivariate covariance matrix, which has an $\mathcal{O}(N_{i}^2)$ complexity. Instead, we rely on the mixture latent variables $\kappa$ and its associated weights $\hat{\mathbf{w}}_{[k]} \in [0,1]^{N_{k}},\; \hat{\mathbf{w}}_{[k]}\geq 0$ and $\sum^{N_{k}}_{\kappa=1} \hat{w}_{\kappa} = 1$ to model the series correlations. We show the relationship between the mixture components and the covariance in Appendix~\ref{appendix:covariance}. Instead of relying on the Markov assumption, we have adopted a joint multi-step forecasting approach that can significantly enhance the computational efficiency of our algorithm. By making predictions in a single forward pass, we can avoid the need for recurrent computations.

In the following subsections, we delve into the properties of the multivariate mixture, such as the analytic version of its implied marginal probability, the relationship between its covariance and the number of mixture components, the bootstrap sample reconciled probability, and the optimization strategies that make it well-suited for large-scale applications. The proofs are inspired on previous work by \citep{olivares2021pmm_cnn}, generalized and extended.

\clearpage
\subsection{Marginal Distributions}~\label{appendix:marginals}

We define the joint distribution of all hierarchical time series in Equation~(\ref{eqn:hierarchical_mixture2}). By integrating the joint probability on the remaining series and time indices, we can obtain the marginal distribution for a single future horizon $\tau \in [t+1:t+h]$ and series $\iota \in [i]$. We express the resulting marginal distribution as follows:
\begin{equation}
\mathbb{P}(Y_{\iota, \tau} = y_{\iota, \tau}|\,\hat{\btheta}) = 
\sum_{\kappa=1}^{N_{k}} w_\kappa 
\mathcal{N} \left(y_{\iota,\tau}|\;\hat{\mu}_{\iota,\kappa,\tau}\;\hat{\sigma}_{\iota,\kappa,\tau} \right)
\end{equation}

\begin{proof}
\begin{equation*}
\begin{split}
    \mathbb{P}(Y_{\iota, \tau} = y_{\iota, \tau}|\,\hat{\btheta}) 
    &= 
    \int^{+\infty}_{-\infty} \ldots \int^{+\infty}_{-\infty}
    \mathbb{P}\left(Y_{[i][t+1:t+h]}=\mathbf{y}_{[i][t+1:t+h]}|\,\hat{\btheta}\right) \delta y_{\iota', \tau'} \setminus \delta y_{\iota, \tau} \\
    & = 
    \int^{+\infty}_{-\infty} \ldots \int^{+\infty}_{-\infty}
    \sum_{\kappa=1}^{N_{k}} \hat{w}_\kappa \; \mathbb{P}(y_{\iota,\tau} \vert \hat{\theta}_{\iota,\kappa,\tau}) \times \prod_{(\iota^\prime,\tau^\prime) \in [i][t+1:t+h] \setminus (\iota, \tau)} \quad \sum_{y_{\iota^\prime,\tau^\prime}}\mathbb{P}(y_{\iota^\prime,\tau^\prime} \vert \hat{\theta}_{\iota^\prime,\kappa,\tau^\prime}) \delta y_{\iota', \tau'} \\
    & = 
    \sum_{\kappa=1}^{N_k} \hat{w}_\kappa \; \mathbb{P}(y_{\iota,\tau} \vert \hat{\theta}_{\iota,\kappa,\tau}) \times 
    \int^{+\infty}_{-\infty} \ldots \int^{+\infty}_{-\infty}
    \prod_{(\iota^\prime,\tau^\prime) \in [i][t+1:t+h] \setminus (\iota, \tau)} \quad \sum_{y_{\iota^\prime,\tau^\prime}}\mathbb{P}(y_{\iota^\prime,\tau^\prime} \vert \hat{\theta}_{\iota^\prime,\kappa,\tau^\prime}) \delta y_{\iota', \tau'}
    \\
    &= \sum_{\kappa=1}^{N_k} w_\kappa \; \mathbb{P}(y_{\iota,\tau} \vert \hat{\theta}_{\iota,\kappa,\tau}) =  \sum_{\kappa=1}^{N_{k}} w_\kappa 
    \mathcal{N} \left(y_{\iota,\tau}|\;\hat{\mu}_{\iota,\kappa,\tau}\;\hat{\sigma}_{\iota,\kappa,\tau} \right)
\end{split}
\end{equation*}
\end{proof}

By removing all other time series and forecast horizons from the joint probability product, the conditional independence expressed in Equation~\ref{eqn:hierarchical_mixture2} can efficiently generate forecast distributions for individual variables.

\subsection{Efficient Covariance Matrix Low-Rank Approximation}~\label{appendix:covariance}

Due to the computational challenges of estimating high-dimensional covariance matrices, existing multivariate methods are limited in their ability to handle a large number of series. To overcome this challenge, our method utilizes a low-rank covariance structure implied by the latent variables of the mixture probability, thereby avoiding the need to compute the covariance matrix explicitly. By doing so, we significantly reduce the number of parameters and enable the modeling of time-varying correlations across millions of time series. 

Let a multivariate random variable $\mathbf{Y}_{[i][t+1:t+h]} \in \mathbb{R}^{N{i}\times h}$ distribution be described by the mixture from Equation~(\ref{eqn:hierarchical_mixture2}), the non-diagonal terms of its implied covariance is the following $N_{k}-1$ rank matrix:
\begin{equation}
\label{eqn:covariance_multivariate}
    \mathrm{Cov}(\mathbf{Y}_{[i],\tau}) = 
    \sum^{N_{k}}_{\kappa=1} \hat{\mathbf{w}}_{\kappa}
    (\hat{\bmu}_{[i],\kappa,\tau}-\bar{\bmu}_{[i],\tau})
    (\hat{\bmu}_{[i],\kappa,\tau}-\bar{\bmu}_{[i],\tau})^{\intercal} 
\end{equation}

\begin{proof}
    We will start by showing that for a pair of series the covariance function is given by:
    \begin{equation}\label{eqn:covariance_pair}
        \mathrm{Cov}(Y_{\iota, \tau}, Y_{\iota', \tau'}) = \overline{\sigma}_{\iota,\tau} \mathbbm{1}(\iota=\iota')\mathbbm{1}(\tau=\tau') + \sum_{\kappa=1}^{N_k} \hat{w}_\kappa \left(\hat{\mu}_{\iota,\kappa,\tau} - \overline{\mu}_{\iota,\tau}\right) \left( \hat{\mu}_{\iota',\kappa,\tau'} - \overline{\mu}_{\iota',\tau'}\right)
    \end{equation}

    By the law of total covariance:
    \begin{equation*}
    \label{eqn:covariance_single}
    \mathrm{Cov}(Y_{\iota, \tau}, Y_{\iota', \tau'}) 
    =    
    \mathbb{E}\left[\mathrm{Cov}(Y_{\iota, \tau}, Y_{\iota', \tau'}\vert \hat{\theta}_{\iota,\kappa,\tau}, \hat{\theta}_{\iota',\kappa,\tau'} ) \right] + 
    \mathrm{Cov}\left(\mathrm{E}\left[Y_{\iota, \tau} \vert \hat{\theta}_{\iota,\kappa,\tau}\right], \mathbb{E}\left[Y_{\iota', \tau'} \vert \hat{\theta}_{\iota',\kappa,\tau'}\right] \right)
    \end{equation*}

    Using the conditional independence from Equation~(\ref{eqn:conditional_indep}). We can rewrite conditional covariance expectation:
    \begin{equation*} \label{eqn:expected_variance}
    \begin{split}
    \mathbb{E}\left[\mathrm{Cov}(Y_{\iota,\tau}, Y_{\iota',\tau'}\vert  \hat{\theta}_{\iota,\kappa,\tau}, \hat{\theta}_{\iota',\kappa,\tau'} ) \right] 
    &= 
    \mathbb{E}\left[\mathrm{Var}(Y_{\iota,\tau}\vert \hat{\theta}_{\iota,\kappa,\tau} ) \right]\mathbbm{1}(\iota=\iota')\mathbbm{1}(\tau=\tau') \\
    &= \mathbb{E}\left[ \hat{\sigma}_{\iota,\kappa,\tau}\right]\mathbbm{1}(\iota=\iota')\mathbbm{1}(\tau=\tau') \\
    &= \overline{\sigma}_{\iota,\tau} \mathbbm{1}(\iota=\iota')\mathbbm{1}(\tau=\tau')
    \end{split}
    \end{equation*}

    where $\overline{\sigma}_{\iota,\tau} = \mathbb{E}\left[ \hat{\sigma}_{\iota,\kappa,\tau}\right] =\sum_{\kappa=1}^{N_k} \hat{w}_\kappa \hat{\sigma}_{\iota, \kappa, \tau} $.

    In the second term, because the conditional distributions are Normal we have 
    
    $$\mathrm{E}\left[Y_{\iota, \tau} \vert \hat{\theta}_{\iota,\kappa,\tau}\right] = \hat{\mu}_{\iota,\kappa,\tau} \quad \text{ and } \quad \mathrm{E}\left[Y_{\iota',\tau'} \vert \hat{\theta}_{\iota',\kappa,\tau'}\right] = \hat{\mu}_{\iota',\kappa,\tau'}$$
    
    Which implies
    
    \begin{equation*}
    \label{eqn:covariance_means}
    \mathrm{Cov}\left(\mathbb{E}\left[Y_{\iota, \tau} \vert \hat{\theta}_{\iota,\kappa,\tau}\right], \mathbb{E}\left[Y_{\iota', \tau'} \vert \hat{\theta}_{\iota',\kappa,\tau'}\right] \right)
    =
    \sum_{\kappa=1}^{N_k} \hat{w}_\kappa 
    \left(\hat{\mu}_{\iota,\kappa,\tau} - \bar{\mu}_{\iota,\tau}\right) 
    \left( \hat{\mu}_{\iota',\kappa,\tau'} - \bar{\mu}_{\iota',\tau} \right)
    \end{equation*}

    Combining the two partial results we recover the pair-wise covariance formula in Equation~(\ref{eqn:covariance_pair}), which can be easily extended to the multivariate case from Equation~(\ref{eqn:covariance_multivariate}). The rank of the matrix can be infered by observing that Equation~(\ref{eqn:covariance_multivariate}) is the sum of $N_{k}$ vectors centered around their means.
\end{proof}

\subsection{Bootstrap Reconciled Probabilities}~\label{appendix:bootstrap}

Let $(\Omega_{[i]}, \mathcal{F}_{[i]}, \hat{\mathbb{P}}(\cdot \,|\, \hat{\btheta}))$ be a probabilistic forecast space, with $\mathcal{F}_{[i]}$ a $\sigma$-algebra on $\mathbb{R}^{N_{i}}$. Let a hierarchical reconciliation transformation be denoted by $\mathbf{SP}(\cdot): \Omega_{[i]} \mapsto \Omega_{[b]} \mapsto \Omega_{[i]}$. Consider $\hat{\mathbf{y}}^{s}_{[i],\tau},\; s=1,\dots,S$ samples drawn from an unconstrained base probability $\hat{\mathbb{P}}(\cdot \,|\, \hat{\btheta})$, and the transformed samples $\tilde{\mathbf{y}}^{s}_{[i],\tau} = \mathbf{SP}\left(\hat{\mathbf{y}}^{s}_{[i],\tau}\right)$.

The probability distribution of the reconciled samples is given by:
\begin{equation}
    \tilde{\mathbb{P}}\left(\tilde{\mathbf{y}}_{[i],\tau}\in\mathcal{H}|\,\tilde{\btheta}\right)
    =
    \hat{\mathbb{P}}\left(\hat{\mathbf{y}}_{[i],\tau} \in \mathbf{SP}^{-1}(\mathcal{H})|\,\hat{\btheta}\right)
\end{equation}

with $\mathcal{H}$ be a coherent forecast measurable set, and $\mathbf{SP}^{-1}(\cdot)$ the reconciliation's inverse image.
\begin{proof} This proof makes only minor modifications to the arguments presented in \cite{panagiotelis2023probabilistic_reconciliation}.
\begin{equation}
\begin{split}
    \tilde{\mathbb{P}}\left(\tilde{\mathbf{y}}_{[i],\tau}\in\mathcal{H}|\,\tilde{\btheta}\right)
    &= \lim_{S \to \infty} \frac{1}{S} \sum^{S}_{s=1} \mathbbm{1}\{\tilde{\mathbf{y}}_{[i],\tau}\in\mathcal{H}\} \\
    &= \lim_{S \to \infty} \frac{1}{S} \sum^{S}_{s=1} \mathbbm{1}\{\hat{\mathbf{y}}_{[i],\tau}\in \mathbf{SP}^{-1}(\mathcal{H})\} \\
    &= \hat{\mathbb{P}}\left(\hat{\mathbf{y}}_{[i],\tau} \in \mathbf{SP}^{-1}(\mathcal{H})|\,\hat{\btheta}\right)
\end{split}
\end{equation}

The first and final equalities follow from the weak law of large numbers, as by definition the indicator functions are independent, identically distributed with a finite mean and variance. The second equality follows from the definition of the inverse image $\mathbf{SP}^{-1}(\mathcal{H})=\{\hat{y}_{[i],\tau}\,|\,\mathbf{SP}(\hat{y}_{[i],\tau}) \in \mathcal{H}\}$.
\end{proof}

A general analytic reconciled probability is derived in Appendix~\ref{appendix:reconciled_probability}. It is worth noting that the bootstrap reconciliation induces a tradeoff between reduced inference speed and the requirement for knowledge of the reconciled parameters $\tilde{\btheta}$.

\clearpage
\subsection{Analytical Reconciled Probabilities}\label{appendix:reconciled_probability}

We use the bootstrap sample reconciliation technique~\citep{panagiotelis2023probabilistic_reconciliation} to ensure the probabilistic coherence of \ours. This technique restores aggregation constraints to base samples, regardless of their distribution. It enhances \ours's modularity by ensuring its probabilistic coherence on a wide range of base probabilities, including non-parametric ones, without requiring any modifications to the original algorithm. We show how a reconciled probability can be recovered analytically through change of variables and marginalization. For simplicity the proofs refer to $\mathbf{y}_{[b]}$ as $\mathbf{b}$, and $\mathbf{y}_{[a]}$ as $\mathbf{a}$.



\textbf{Lemma.} Consider the classic reconciliation approach where the entire hierarchy's forecasts are combined into reconciled bottom-level forecasts using a composition of linear transformations $\mathbf{SP}(\cdot)=\mathbf{S_{[i][b]}P_{[b][i]}}(\cdot)$. The reconciled probability for the new bottom-level series is given by:
\begin{equation}\label{eqn:reconciled_bottom}
\tilde{\mathbb{P}}_{[b]}\left(\mathbf{b}\right) = |\mathbf{P}^{*}| \int \hat{\mathbb{P}}_{[i]}\left(\mathbf{P}_{\perp} \mathbf{a} +  \mathbf{P}^{-} \mathbf{b}\right) \delta \mathbf{a}
\end{equation}
where $\hat{\mathbb{P}}\left(\cdot\right)$ is the unconstrained base forecast distribution, $\mathbf{P}_{\perp} \in \mathbb{R}^{N_{i} \times N_{a}}$, $\mathbf{P}^{-}  \in \mathbb{R}^{N_{i} \times N_{b}}$ are the orthogonal complement of $\mathbf{P}_{[i][b]}$ and its Moore-Penrose inverse; matrix $\mathbf{P}^{*} = \left[\mathbf{P}_{\perp} \,|\, \mathbf{P}^{-} \right]$, and bottom level $\mathbf{b}$ and aggregate level $\mathbf{a}$ vectors are obtained through the following variable change:
\begin{equation}
\hat{\mathbf{y}}_{[i]}
= \mathbf{P}^{*}
\begin{bmatrix}
\mathbf{a} \\
\mathbf{b} 
\end{bmatrix}
\end{equation}

\begin{proof} Using the multivariate change of variables theorem, and properties of the Jacobian of a linear mapping:
\begin{equation}
\tilde{\mathbb{P}}(\mathbf{a},\mathbf{b})
=
\left|\det \left[\left.{\frac {d\mathbf{P}^{*}(\mathbf{z} )}{d\mathbf{z} }}\right|_{\mathbf{z} =(\mathbf{a},\mathbf{b})}\right]\right|
\mathbb{P}{\Bigl (}\mathbf{P}^{*} 
\begin{bmatrix}
\mathbf{a}
\mathbf{b} 
\end{bmatrix}
{\Bigr )} 
= |\mathbf{P}^{*}| \mathbb{P}_{[i]}\left(\mathbf{P}_{\perp} \mathbf{a} +  \mathbf{P}^{-} \mathbf{b}\right) \qquad \qquad \qquad \qquad
\end{equation}

Marginalizing $\mathbf{a}$ we obtain the reconciled probability for the bottom level series.
\end{proof}

\textbf{Theorem 3.1:} Consider the classic reconciliation approach where the entire hierarchy's forecasts are combined into reconciled bottom-level forecasts using a composition of linear transformations $\mathbf{SP}(\cdot)=\mathbf{S_{[i][b]}P_{[b][i]}}(\cdot)$. The reconciled probability for the entire hierarchical series is given by:
\begin{equation}\label{eqn:reconciled_hierarchical}
    \tilde{\mathbb{P}}\left(\mathbf{y}_{[i]}\right)
    =
    |\mathbf{S}^{*}|
    \tilde{\mathbb{P}}_{[b]}\left(\mathbf{S}^{-} \mathbf{y}_{[i]} \right) \mathbbm{1}\{\mathbf{y}_{[i]} \in \mathcal{H}\}
\end{equation}

where $\tilde{\mathbb{P}}_{[b]}(\cdot)$ is the reconciled bottom forecast distribution, $\mathbbm{1}(\tilde{\mathbf{y}}_{[i]} \in \mathcal{H})$ indicates if realization belongs in the $N_{b}$-dimensional hierarchically coherent subspace $\mathcal{H}$, $\mathbf{S}^{-} \in \mathbb{R}^{N_{b} \times N_{i}}$ is $\mathbf{S}_{[i][b]}$ Moore-Penrose inverse and $\mathbf{S}_{\perp} \in \mathbb{R}^{N_{i} \times N_{a}}$ its orthogonal complement.

\begin{proof} This proof follows closely that provided in \cite{panagiotelis2023probabilistic_reconciliation}. Given bottom level forecast distribution from the Lemma, one can create a degenerate distribution for the entire hierarchy by adding additional dimensions $\mathbf{u} \in \mathbb{R}^{N_{a}}$.
\begin{equation}
    \tilde{\mathbb{P}}_{[i]}(\mathbf{u},\mathbf{b}) = \tilde{\mathbb{P}}_{[b]}\left(\mathbf{b}\right) \mathbbm{1}\{\mathbf{u} = \mathbf{0}\}
\end{equation}
Let $\mathbf{S}=\mathbf{S}_{[i][b]}$ and $\mathbf{S}^{\intercal}_{\perp}$ its orthogonal complement, and $\mathbf{S}^{-}$ and $\mathbf{S}^{-}_{\perp}$ the respective Moore-Penrose inverses. Using the following change of variables $\mathbf{b}=\mathbf{S}^{-}\mathbf{y}_{[i]}$ and $\mathbf{u}=\mathbf{S}^{\intercal}_{\perp}\mathbf{y}_{[i]}$ we obtain
\begin{equation}
\mathbf{y}_{[i]}
= \left[\mathbf{S}^{-}_{\perp} \,|\, \mathbf{S} \right]
\begin{bmatrix}
\mathbf{u} \\
\mathbf{b} 
\end{bmatrix}
\quad \iff \quad
\mathbf{S}^{*} \mathbf{y}_{[i]}
=
\begin{bmatrix}
\mathbf{S}^{\intercal}_{\perp} \\
\mathbf{S}^{-} 
\end{bmatrix}
\mathbf{y}_{[i]}
= 
\begin{bmatrix}
\mathbf{u} \\
\mathbf{b} 
\end{bmatrix}
\end{equation}

\begin{equation}
    \tilde{\mathbb{P}}\left(\mathbf{y}_{[i]}\right)
    =
    |\mathbf{S}^{*}|
    \tilde{\mathbb{P}}_{[b]}\left(
    \mathbf{S}^{\intercal}_{\perp}\mathbf{0} + 
    \mathbf{S}^{-} \mathbf{y}_{[i]} \right) 
    \mathbbm{1}\{\mathbf{S}^{\intercal}_{\perp}\mathbf{y}_{[i]} = \mathbf{0} \}
    =
    |\mathbf{S}^{*}|
    \tilde{\mathbb{P}}_{[b]}\left(\mathbf{S}^{-} \mathbf{y}_{[i]} \right) 
    \mathbbm{1}\{\mathbf{y}_{[i]} \in \mathcal{H}\}
\end{equation}

By definition of the orthogonal complement if $\mathbf{S}^{\intercal}_{\perp}\mathbf{y}_{[i]} = \mathbf{0}$, that means that $\mathbf{y}_{[i]} \in \mathrm{span}(\mathbf{S})$, that matches the definition of the hierarchically coherent subspace $\mathcal{H}$.

\end{proof}

As mentioned earlier, the analytical version of reconciled probabilities can provide highly efficient inference times, depending on the properties of the reconciliation and the base forecast distributions. For instance, recent studies have utilized Gaussian distributions~\citep{panagiotelis2023probabilistic_reconciliation, wickramasuriya2023probabilistic_gaussian}, and Poisson Mixtures~\citep{olivares2022hierarchicalforecast}.


\subsection{\ours\ Parameter Estimation}

\textbf{Maximum Likelihood Estimation}

To estimate the \ours's parameters, we can use maximum likelihood estimation for the multivariate probability in Equation~(\ref{eqn:hierarchical_mixture}). Let $\bomega$ denote \ours's parameters that condition the probabilistic output layer parameters. Then, we express the negative log-likelihood function as follows:
\begin{equation}\label{eq:negative_log_joint_likelihood}
    \mathcal{L}(\bomega) = - \mathrm{log} \Bigg[\sum_{\kappa=1}^{N_{k}} \hat{w}_\kappa(\bomega) \prod_{(\iota,\tau) \in [i][t+1:t+h]} 
        \left( 
        \frac{1}{\hat{\sigma}_{\iota,\kappa,\tau}(\bomega) \sqrt{2 \pi}}
    \exp{ \bigl\{
    -\frac{1}{2}
    \left(
    \frac{y_{\iota,\kappa,\tau}-\hat{\mu}_{\iota,\kappa,\tau}(\bomega)}{ \hat{\sigma}_{\iota,\kappa,\tau}(\bomega) } 
    \right)^{2}
    \bigl\} }
    \right) \Bigg]
\end{equation}

Although standard maximum likelihood estimation can model relationships between multiple time series across the forecast horizon, a scalability challenge arises when the number of series and forecast horizon increase significantly. Since its computation requires access to the entire multivariate series, MLE can become computationally intensive and time-consuming. 

\textbf{Maximum Composite Likelihood Estimation}

Composite likelihood provides a computationally efficient alternative to maximum likelihood estimation for optimizing the parameters of \ours. Unlike MLE, which computes the whole multivariate likelihood, composite likelihood decomposes the hierarchical variable high-dimensional space support into sub-spaces and optimizes the weighted product of the subspaces' marginal likelihood. When defining the sub-spaces in composite likelihood, the probabilistic model is restricted to learning relationships within each sub-space while assuming independence across non-overlapping sub-spaces. These sub-spaces can be defined based on the user's application needs. For instance, they can be guided by the geographic proximity of the time series data. In order to simplify the \ours\ algorithm, we randomly assign each series to the sub-spaces defined by the stochastic gradient batches. Let $\mathcal{B}=\{[b_i]\}$ be time-series SGD batches, then \ours\'s negative log composite likelihood is:
\begin{equation}\label{eqn:composite_likelihood}
    \mathcal{L}(\bomega) = - \sum_{[b_{i}] \in \mathcal{B}}
	\mathrm{log} \left[\sum_{\kappa=1}^{N_{k}} \hat{w}_\kappa(\bomega) \prod_{(\iota,\tau) \in [b_{i}][t+1:t+h]}
	    \left( 
	    \frac{1}{\hat{\sigma}_{\iota,\kappa,\tau}(\bomega) \sqrt{2 \pi}}
	    \exp{ \bigl\{
	    -\frac{1}{2}
	    \left(
	    \frac{y_{\iota,\kappa,\tau}-\hat{\mu}_{\iota,\kappa,\tau}(\bomega)}{ \hat{\sigma}_{\iota,\kappa,\tau}(\bomega) } 
	    \right)^{2}
	    \bigl\} }
	    \right) \right]
\end{equation}

Composite likelihood's independent sub-spaces can be defined as each series forecast, leading to the univariate estimation approach.
\begin{equation}\label{eq:composite_likelihood_univariate}
    \mathcal{L}_{univ}(\bomega) = - \sum_{\iota \in [i]}
	\mathrm{log} \left[\sum_{\kappa=1}^{N_{k}} \hat{w}_\kappa(\bomega) \prod_{\tau \in [t+1:t+h]}
	    \left( 
	    \frac{1}{\hat{\sigma}_{\iota,\kappa,\tau}(\bomega) \sqrt{2 \pi}}
	    \exp{ \bigl\{
	    -\frac{1}{2}
	    \left(
	    \frac{y_{\iota,\kappa,\tau}-\hat{\mu}_{\iota,\kappa,\tau}(\bomega)}{ \hat{\sigma}_{\iota,\kappa,\tau}(\bomega) } 
	    \right)^{2}
	    \bigl\} }
	    \right) \right]
\end{equation}

Figure~\ref{fig:forecast_comparison} compares univariate and composite likelihood estimation.

\clearpage
\section{Hierarchical Dataset's Exploration}
\label{section:hierarchical_datasets}
In this Appendix we complement the description of the benchmark datasets from Section~\ref{section4:evaluation}.

\begin{table}[htpb] 
\fontsize{4}{4}\selectfont
    \vskip -0.1in
    \caption{Summary, of experiment hierarchical datasets.}
    \begin{center}
    \begin{small}
    \label{table:datasets_summary}
	\begin{sc}
            \setlength\tabcolsep{1.4pt}
		\begin{tabular}{lcccccc}
			\toprule
			Dataset        & Total   & Aggregate     & Bottom          & Frequency.   & h    & Levels \\ \midrule   
			\Labour        & 57      & 25            & 32              & Month        &  8   & 4      \\            
                \Traffic       & 207     & 7             & 200             & Daily        &  7   & 3      \\            
			\Tourism       & 89      & 33            & 56              & Quarterly    &  4   & 4      \\            
			\TourismL      & 555     & 175           & $76 \,/\, 304$  & Month        &  12  & 4/5    \\            
                \Wikitwo       & 199     & 49            & 150             & Daily        &  7   & 5      \\ \bottomrule
		\end{tabular}
	\end{sc}
        \end{small}
	\end{center}
	\vskip -0.1in
\end{table}

\begin{figure}[ht]
    \centering
    \subfigure[Labour]{
    \label{fig:labour_agg_matrix}
    \includegraphics[width=0.17\linewidth]{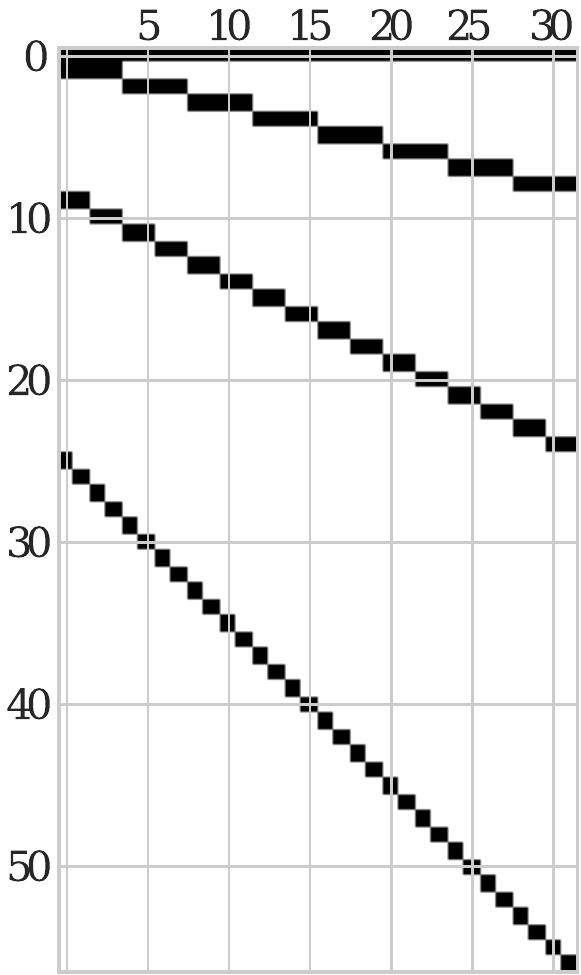}
    }
    \subfigure[Traffic]{
    \label{fig:traffic_agg_matrix}
    \includegraphics[width=0.17\linewidth]{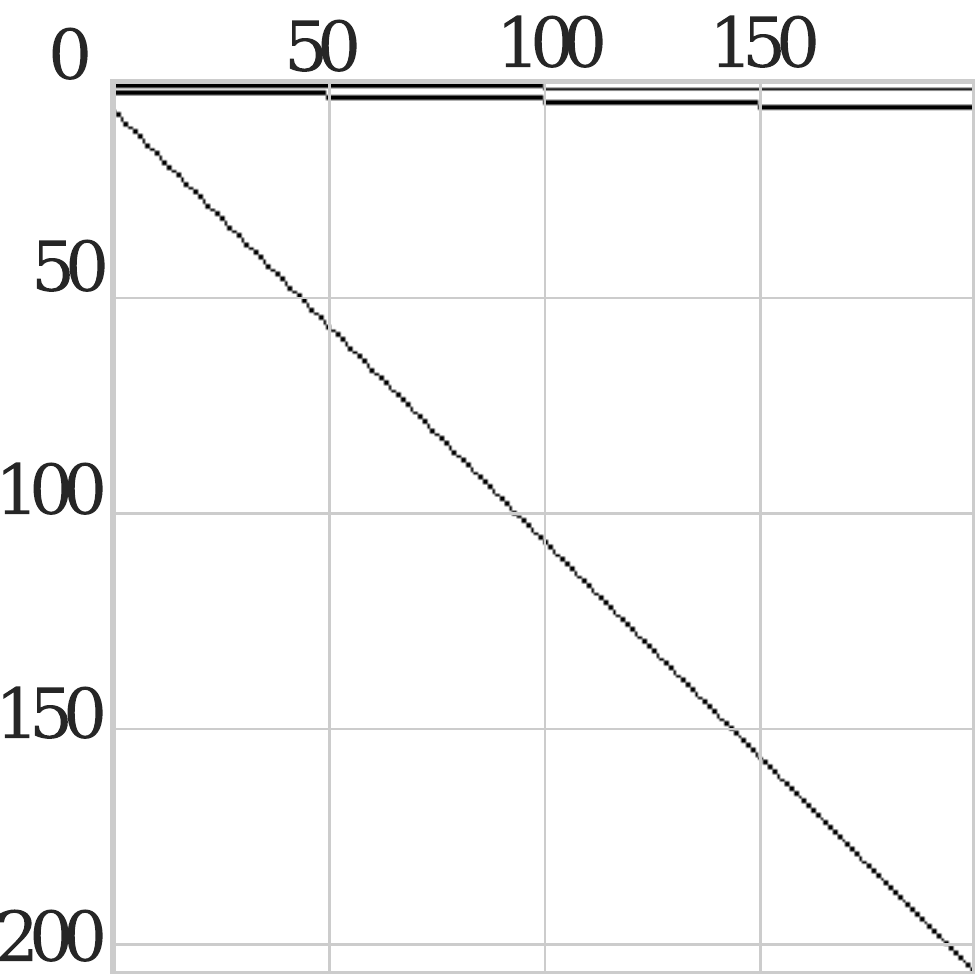}
    }
    \subfigure[TourismS]{
    \label{fig:tourism_small_agg_matrix}
    \includegraphics[width=0.17\linewidth]{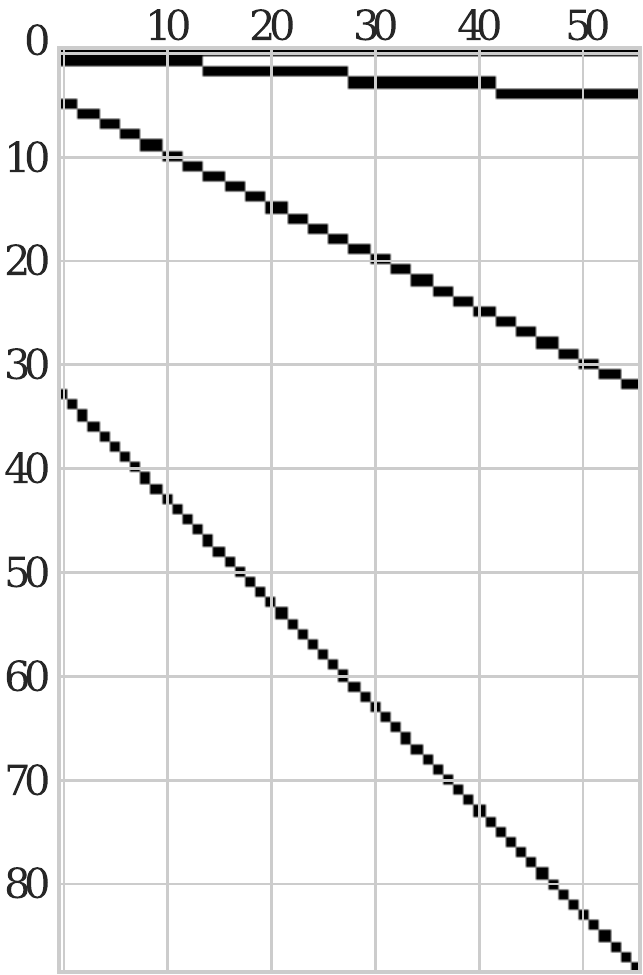}
    }
    \bigskip
    \subfigure[TourismL]{
    \label{fig:tourism_large_agg_matrix}
    \includegraphics[width=0.17\linewidth]{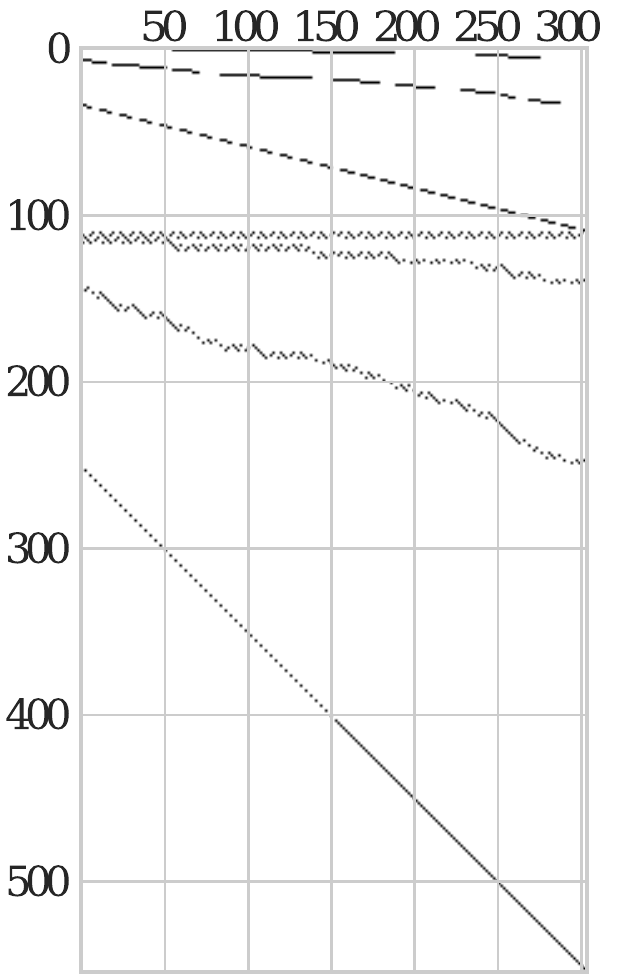}
    }
    \subfigure[Wiki2]{
    \label{fig:wiki_agg_matrix}
    \includegraphics[width=0.17\linewidth]{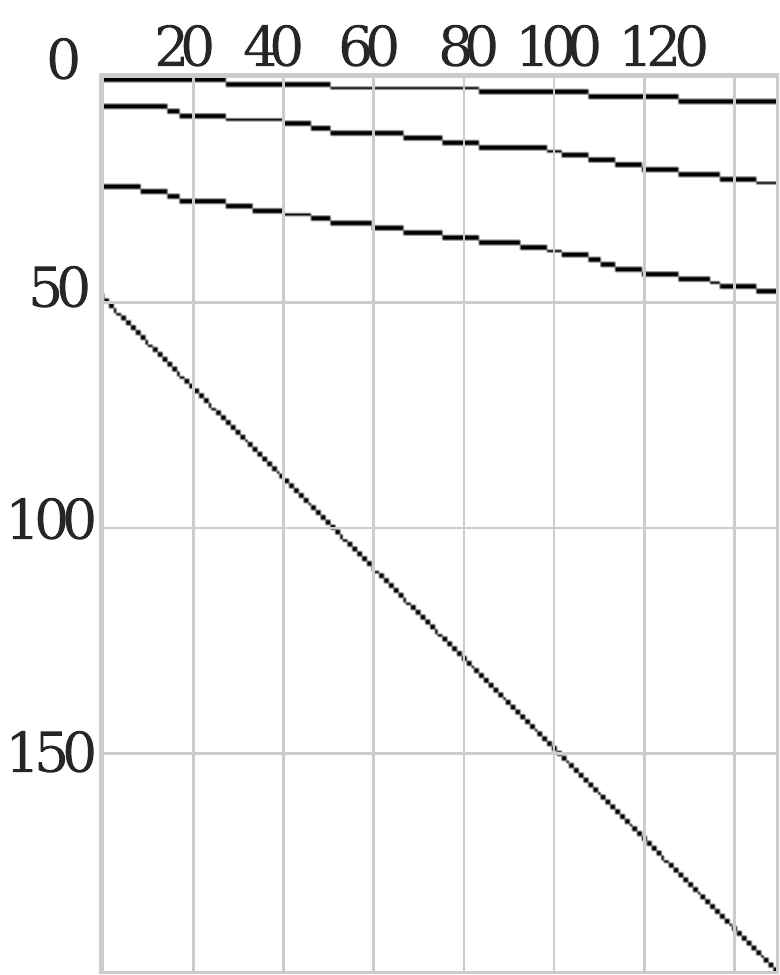}
    }
    \caption{Dataset's hierarchical constraints. (a) \Labour\ groups 32 occupation series by gender and geography. (b) \Traffic\ groups 200 highways' occupancy series into quarters, halves and total. (c) \Tourism\ groups 56 quarterly Australian tourist visits by geographic levels. (d) \TourismL\ groups its 555 monthly Australian regional visit series, into a combination travel purpose, zones, states and country geographical aggregations. (e) \Wikitwo\ groups 150 daily visits to Wikipedia articles by language and article categorical taxonomy.}
    \label{fig:aggregation_matrices}
\end{figure}

\begin{figure}[ht] 
\centering     
\includegraphics[height=15.00 cm]{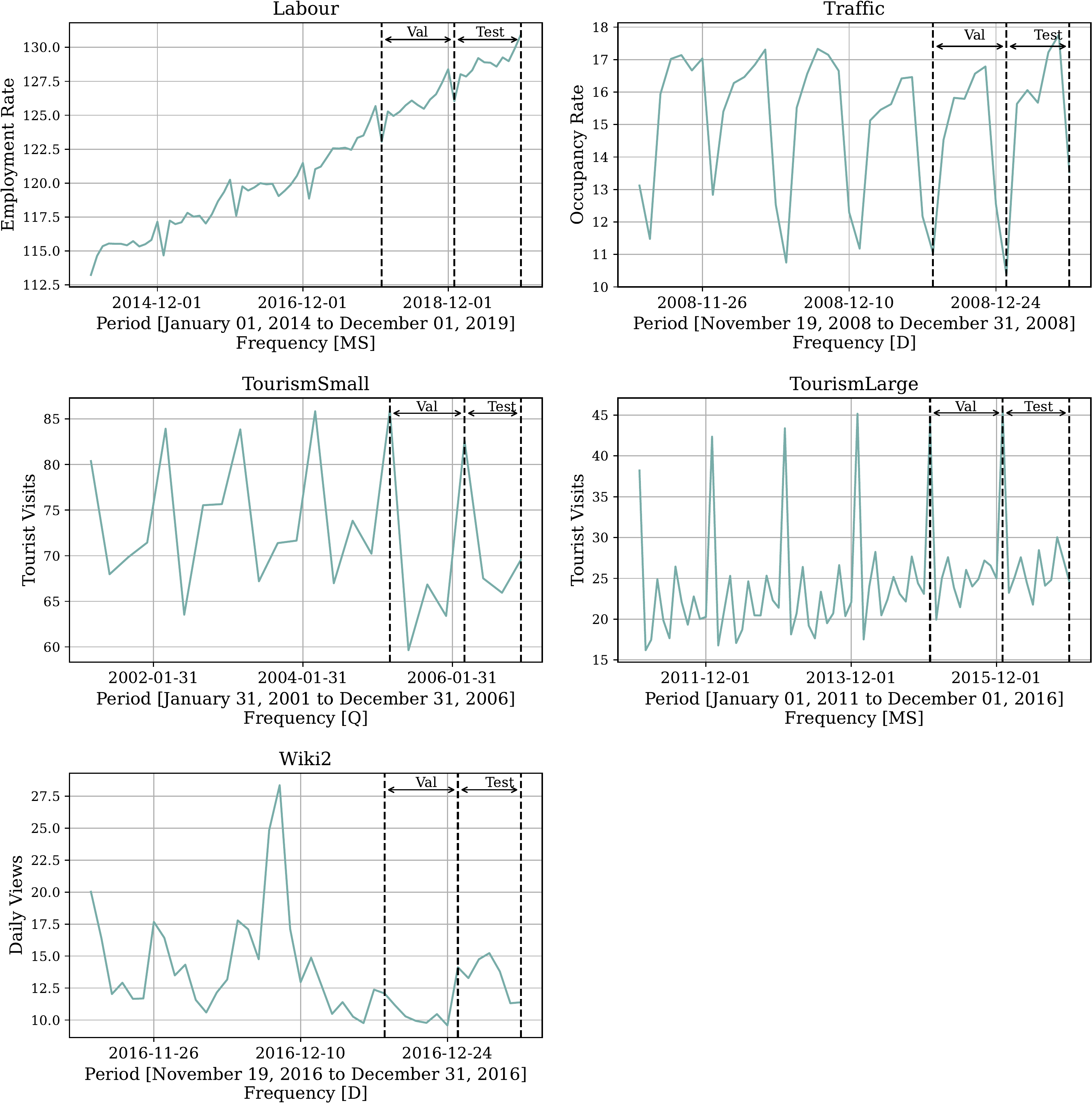}
\caption{Datasets' partition into train, validation, and test sets used in our experiments. All use the last horizon window as defined in Table~\ref{table:datasets_summary} (marked by the second dotted line), and the previous window preceding the test set as validation (between the first and second dotted lines). Validation  provides the signal for hyperparameter optimization.}
\label{fig:train_methodology}
\end{figure}

\vspace{1cm}
\Labour~ reports monthly Australian employment from February 1978 to December 2020. It contains a structure built by the labour categories \citep{Aulabor2019Aulabor_dataset}.
\Traffic~ measures the occupancy of 963 traffic lanes in the Bay Area, the data is grouped into a year of daily observations and organized into a 207 hierarchical structure \citep{dua2017traffic_dataset}.
\Tourism~ consists of 89 Australian location quarterly visits series; it covers from 1998 to 2006. Several studies have used this dataset in the past \citep{canberra2005tourismS_dataset}.
\TourismL~ summarizes an Australian visitor survey managed by the Tourism Research Australia, the dataset contains 555 monthly series from 1998 to 2016, and it is organized into geographic and purpose of travel \citep{canberra2019tourismL_dataset}.
\Wikitwo~ contains the daily views of 145,000 Wikipedia articles from July 2015 to December 2016. The dataset is filtered and processed into 150 bottom series and 49 aggregate series \citep{wikipedia2018web_traffic_dataset}. Figure~\ref{fig:train_methodology} shows each dataset's most aggregated series along with its training methodology partition. Figure~\ref{fig:aggregation_matrices} shows each dataset's hierarchical aggregation constraint matrices.

\vspace{1cm}

\clearpage
\section{Ablation Studies} 
\label{section:ablation_studies}
In this Appendix, we perform ablation studies on the validation set of five hierarchical datasets \Labour, \Traffic, \Tourism, \TourismL, and \Wikitwo.
For these experiments, we change minimally the \ours\ settings defined in Table~\ref{table:hyperparameters}, comparing normalization techniques, varying the number of Mixture components, and exploring different hierarchical reconciliation strategies to understand their contribution to the performance of the method.

\subsection{Scaled Decoupled Optimization}
\label{section:scaled_decoupled_ablation}


Section~\ref{section:scale_decouple}, introduced \ours's scale decouple optimization strategy with the \TemporalNorm\ transform. Here we study the effects of different temporal normalization strategies on the forecast accuracy performance of the model, measured with the overall sCRPS. For simplicity consider a network with only temporal input $\mathbf{x}^{h}_{[i][:t][c]}$, with $[i]$ batch, $[:t]$ time, and $[c]$ feature channel indexes, we consider transformations of the general form:
\begin{equation}
\label{eqn:temporal_normalization2}
\begin{split}
    \check{\mathbf{x}}^{h}_{[i][:t+h][c]}
    & = \mathrm{TemporalNorm}(\mathbf{x}^{h}_{[i][:t+h][c]}) = \frac{\mathbf{x}^{h}_{[i][:t][c]}-\mathbf{a}}{\mathbf{b}} \\
    \hat{\theta}(\check{\mathbf{x}}^{h}_{[i][:t+h][c]}) & = \mathrm{TemporalNorm}^{-1}(\bomega(\check{\mathbf{x}}_{[i][:t+h][c]})) = \mathbf{b} \bomega_{[i][t+h]} + \mathbf{a} 
\end{split}
\end{equation}

where $\mathbf{a},\mathbf{b} \in \mathbb{R}^{(N_{a}+N_{b}) \times N_{c}}$ is the shift and the scale of the historic inputs. In this experiment we augment \NHITS, \Autoformer, \LSTM, \MLP, and \TFT\ with three different \TemporalNorm\ normalization schemes, along with the recently proposed Reversible Instance Normalization (RevIN) technique:

\begin{gather*}
\label{eqn:normalizers}
    \begin{alignedat}{2}
    \mathrm{minmax}   &: 
    \frac{(\mathbf{x}^{h}_{[i][:t][c]}-\mathrm{min}(\mathbf{x}^{h}_{[i][:t][c]})_{[i][c]})}{(\mathrm{max}(\mathbf{x}^{h}_{[i][:t][c]})_{[i][c]}- \mathrm{min}(\mathbf{x}^{h}_{[i][:t][c]})_{[i][c]}) }
    &\quad\quad 
    \mathrm{standard} &: \frac{(\mathbf{x}^{h}_{[i][:t][c]}-\bar{\mathbf{X}}_{[i][c]})}{\hat{\sigma}_{[i][c]}} \\
    && \\
    \mathrm{robust}   &: \frac{(\mathbf{x}^{h}_{[i][:t][c]}-\mathrm{median}(\mathbf{x}^{h}_{[i][:t][c]})}{\mathrm{mad}(\mathbf{x}^{h}_{[i][:t][c]})_{[i][c]}}
    &\quad\quad
    \mathrm{revin} &: \lambda_{[i][c]} * \frac{(\mathbf{x}^{h}_{[i][:t][c]}-\bar{\mathbf{X}}_{[i][c]})}{\hat{\sigma}_{[i][c]}} + \beta_{[i][c]}\\
    \end{alignedat} 
\end{gather*}

Figure 4 compares the four different normalization techniques across different architectures for different levels of noise, on the \TourismL \, dataset. It is evident that the performance improvements vary depending on the type of temporal normalization employed. In cases where signals are smooth, the type of normalization used makes little difference. However, as the signal becomes noisier and more volatile, the robust normalization strategy consistently outperforms all other normalization techniques, with the minmax strategy lagging behind. This underscores the advantages of employing median shift and mad scale to enhance the model's robustness against large outliers in the data. As a result, we have selected robust Temporal Normalization as the default option for all subsequent experiments.

\begin{figure*}[ht]
\label{fig:scale_decouple_optimization}
\centering
\begin{tabular}{ccc}
    \subfigure[Autoformer]{\includegraphics[width=0.21\textwidth]{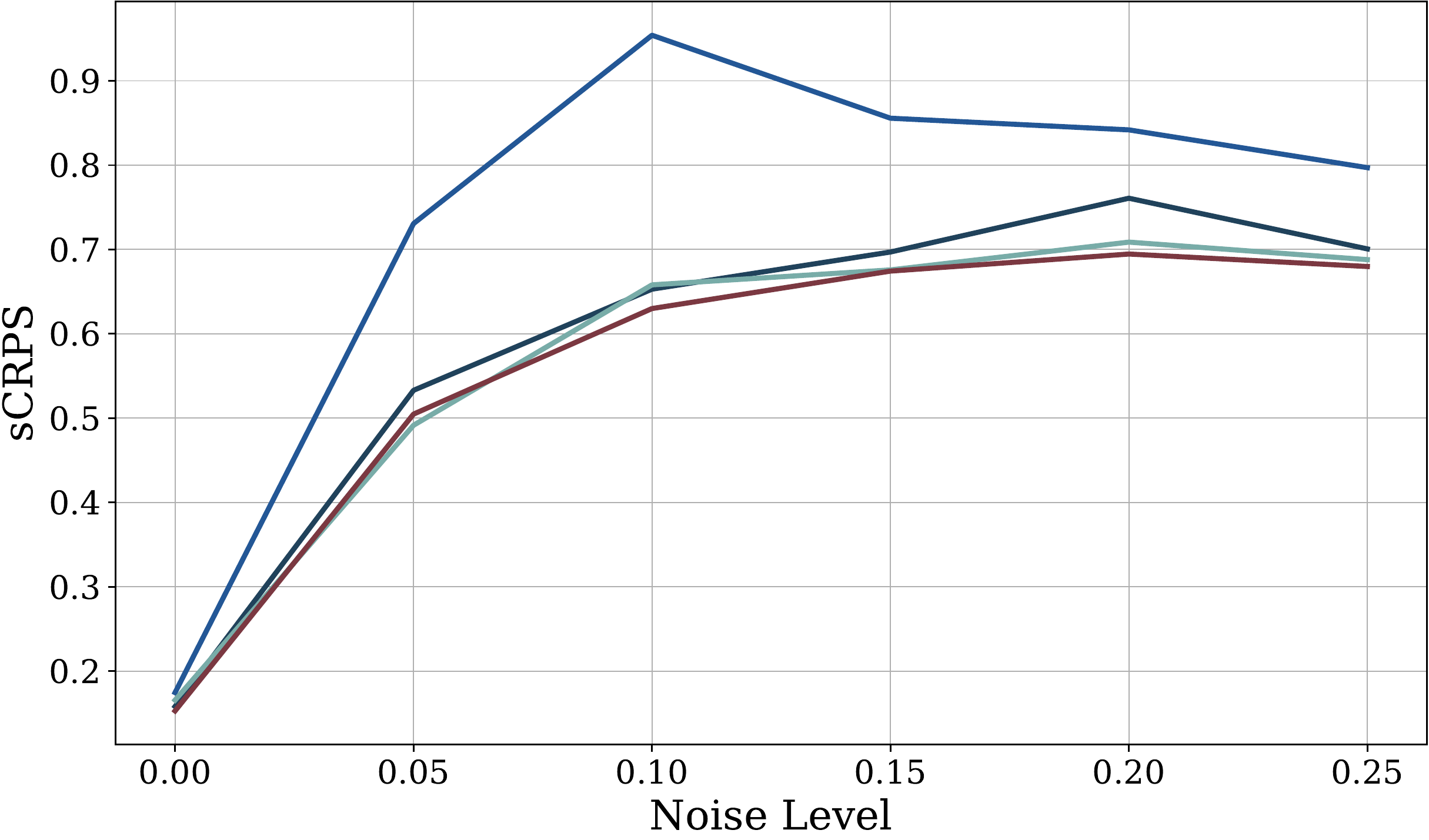}} 
    &
    \subfigure[LSTM]{\includegraphics[width=0.21\textwidth]{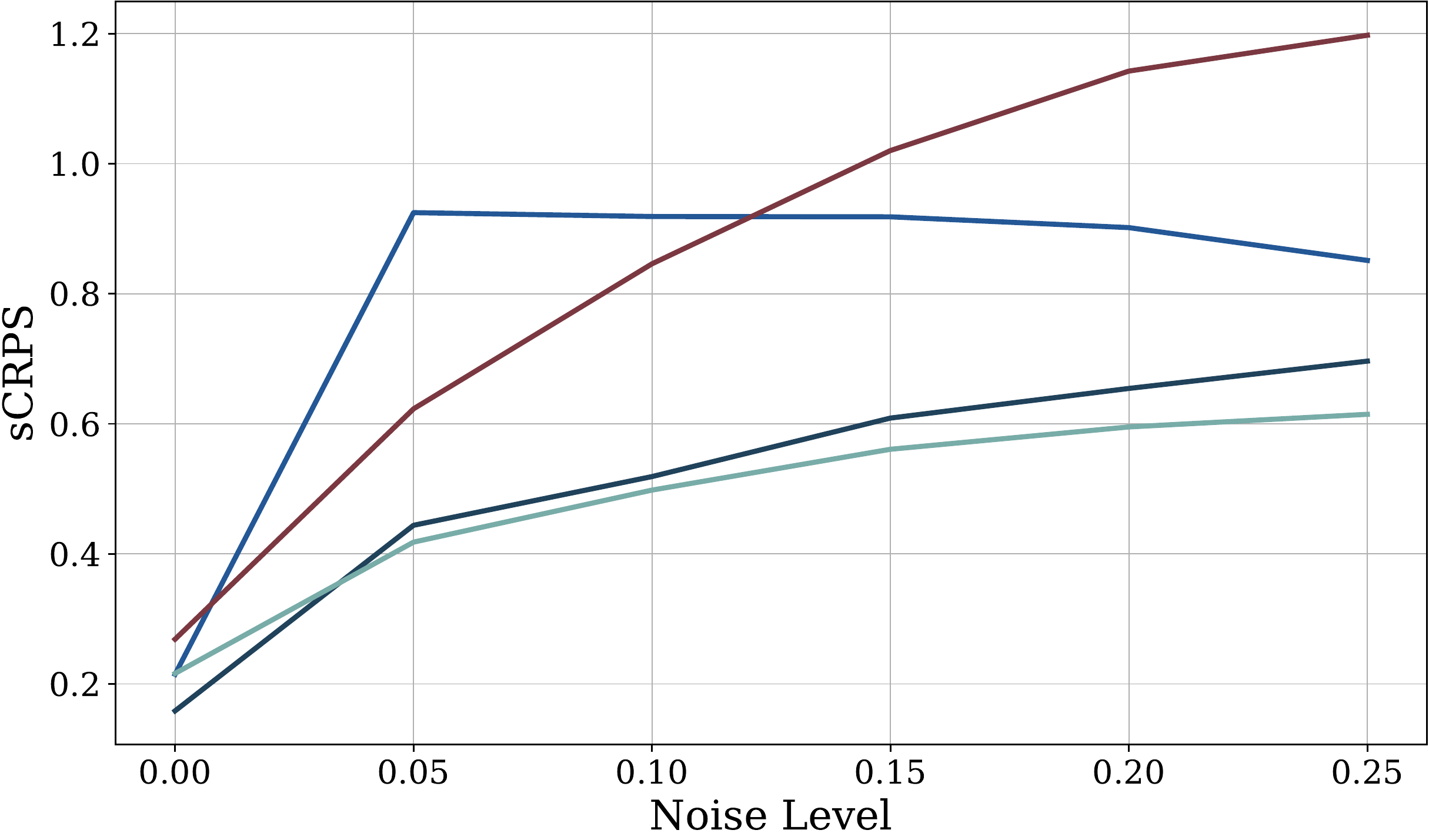}} 
    &
    \subfigure{\includegraphics[width=0.21\textwidth]
    {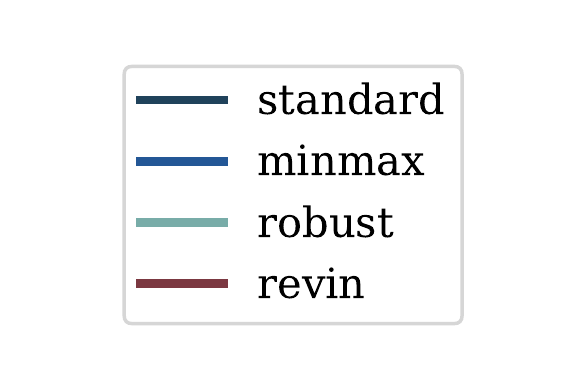}} \\
    \subfigure[MLP]{\includegraphics[width=0.21\textwidth]{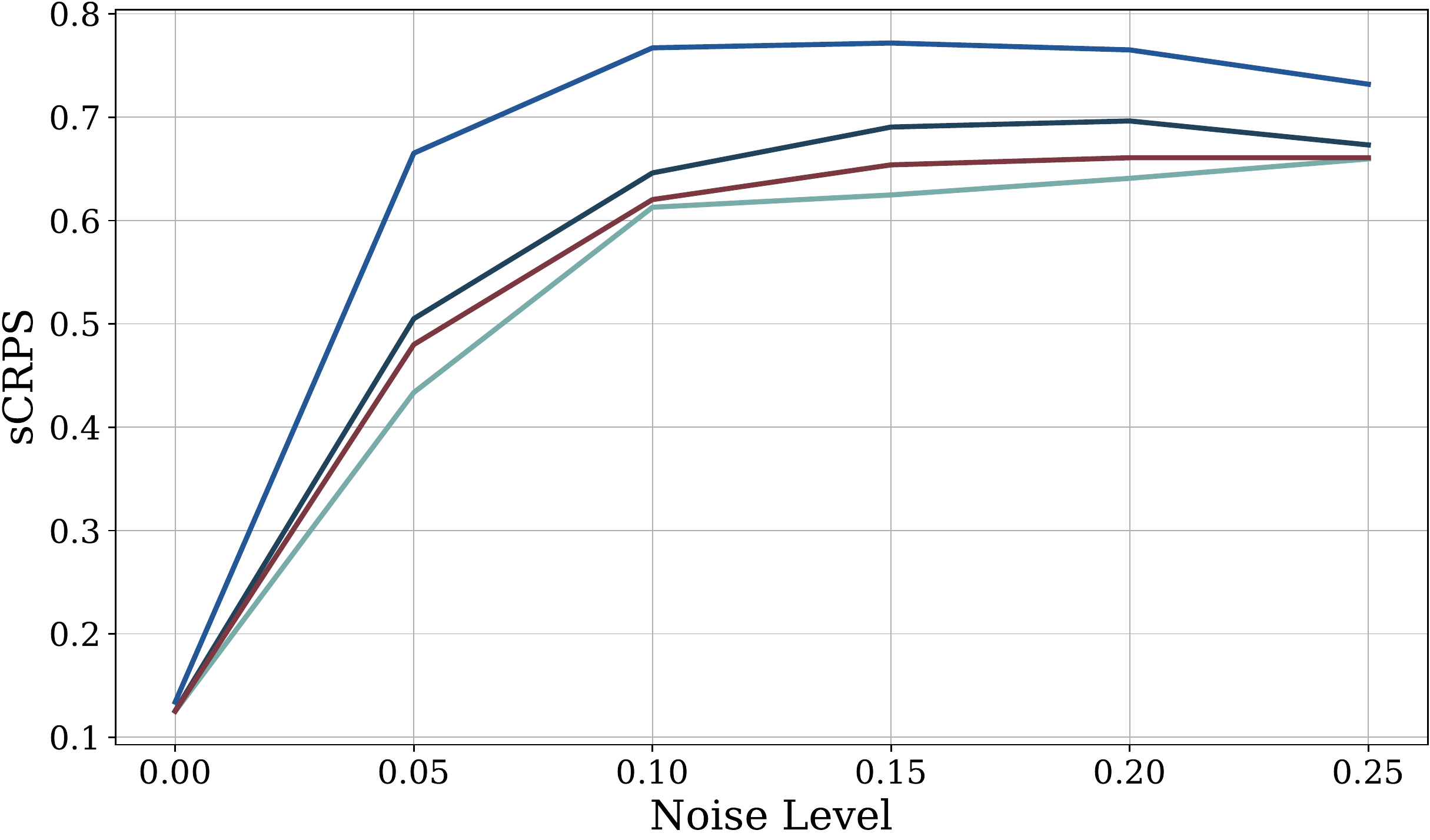}} 
    &
    \subfigure[NHITS]{\includegraphics[width=0.21\textwidth]{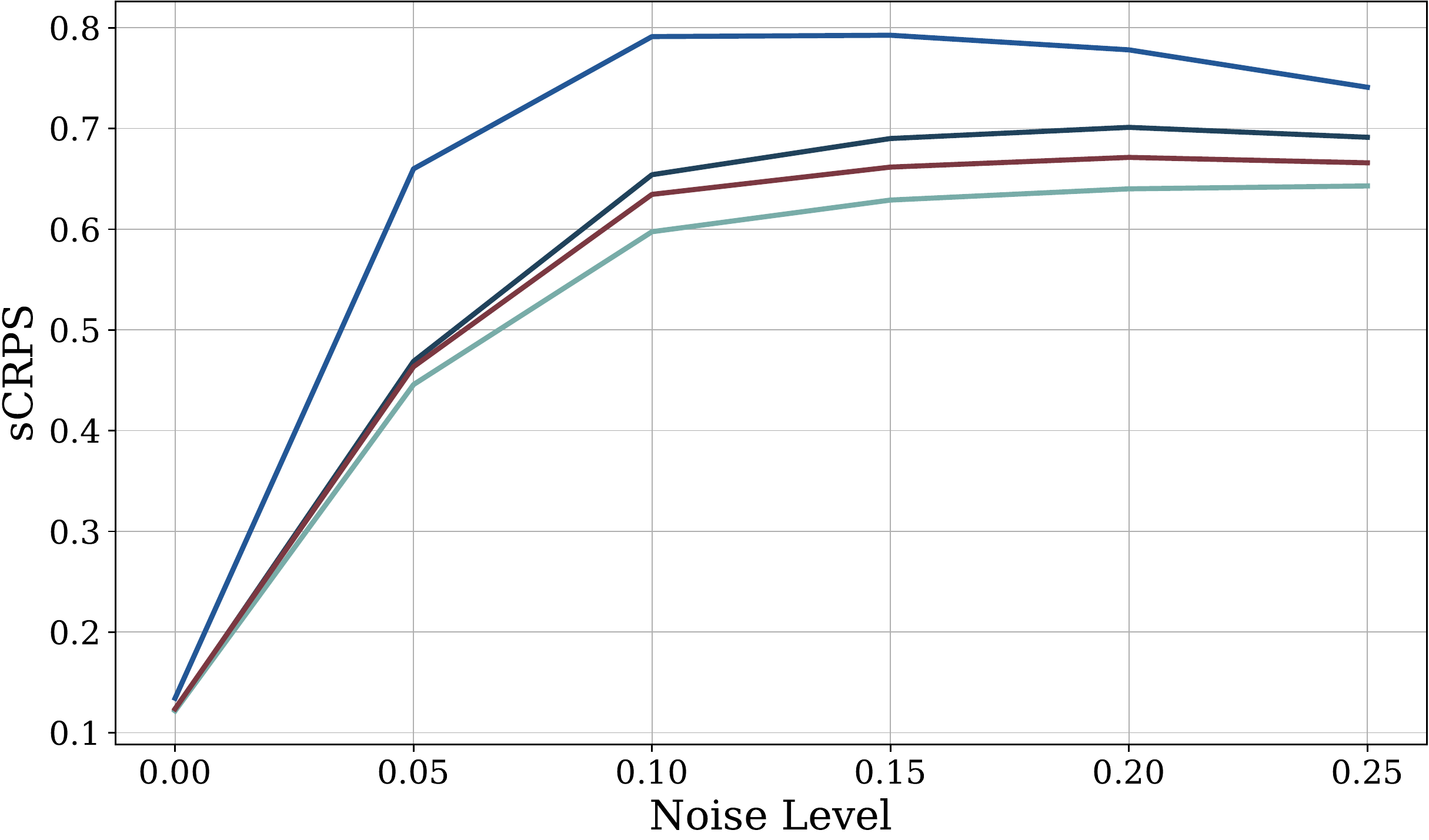}} 
    &
    \subfigure[TFT]{\includegraphics[width=0.21\textwidth]{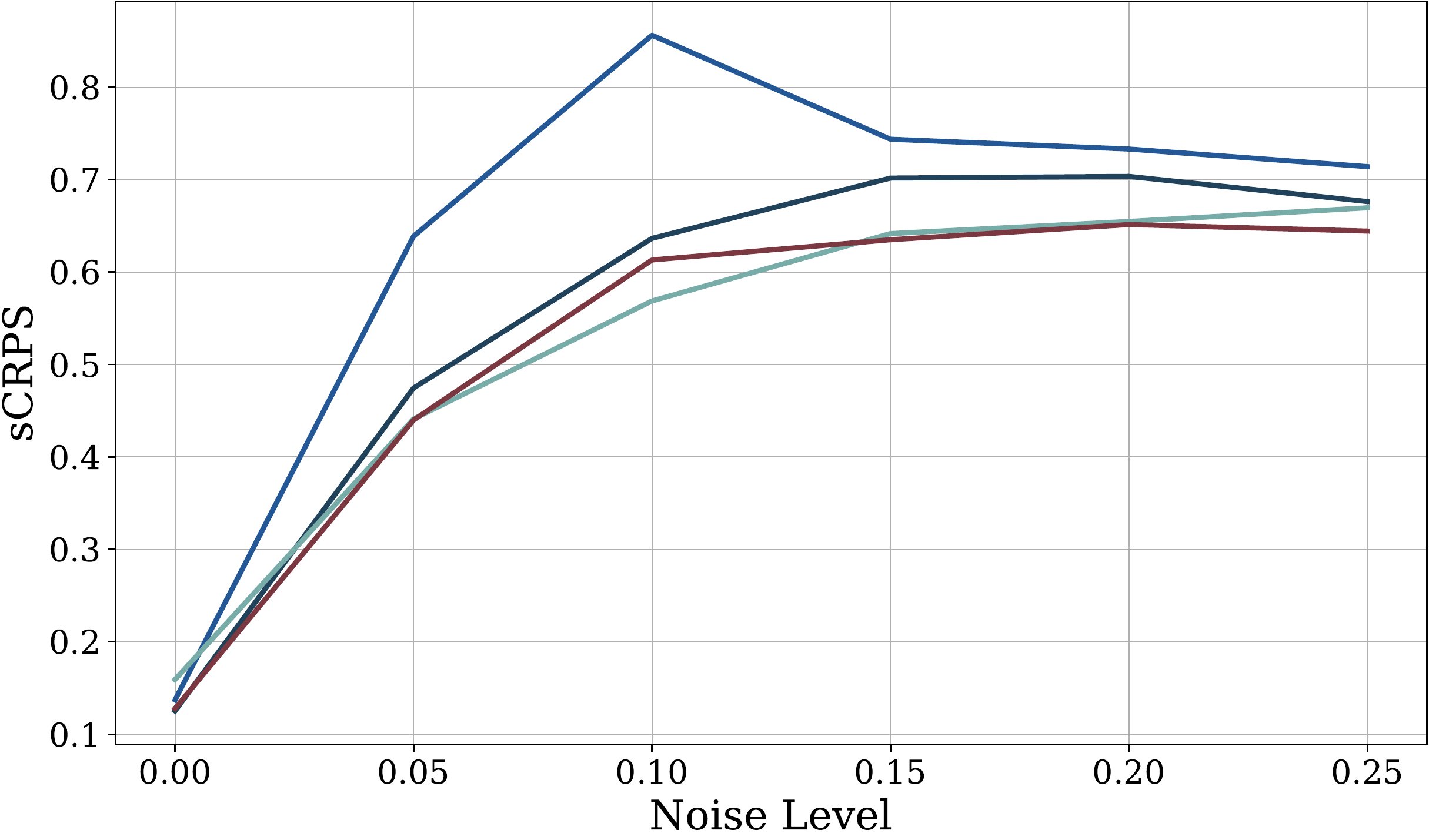}}
\end{tabular}
\caption{Validation \emph{scaled Continuous Ranked Probability Score} (sCRPS) curves on \TourismL \, for five different architectures and varying noise levels. For each noise level \emph{p}, we transformed a fraction \emph{p} of the training dataset by a random scaling factor. We see the robust normalization (green curve) is the lowest for all five architectures, showing how it consistently outperforms the other normalization techniques at larger noise levels.}
\end{figure*}

\clearpage
\subsection{Mixture Size Exploration}
\label{section:mixture_ablation}

As mentioned in Section~\ref{section:multivariate_mixture} and proven in Appendix~\ref{section:hierarchical_mixture_properties}, our multivariate mixture probability model is capable of capturing the relationships among the hierarchical series and its expressivity directly determined by the number of components in the Gaussian mixture. Here we conduct a study to compare the accuracy effects of different mixture sizes. We vary the number of components monotonically and follow the model's performance. An \NHITS\ model configuration is automatically selected as defined from Table~\ref{table:hyperparameters}.

During our \Labour, \Traffic, and \TourismL\ experiments, we observed a bias-variance trade-off relationship between the number of mixture components and the validation sCRPS. If we set the number of components to 1, the sCRPS score is the highest, but as the number of components increases, the sCRPS improves. However, if there are too many components, the CRPS score worsens again after a certain point. Our theory is that if the mixture has too few components, the probability does not have enough parameters to accurately depict the data's correlations; this leads to a high bias and reflects in a poor sCRPS. On the other hand, if the mixture has too many components, the model becomes too complicated and quickly overfits the training data, resulting in high variance and poor performance on the validation data. The results of this ablation experiment show a clear benefit of using a flexible multivariate mixture distribution, contrasting it to the simpler approach of using a single component (Gaussian/Poisson regression). We explain these improvements from the flexibility of the mixture approach that operates as a Kernel density estimation and can arbitrarily approximate a wide variety of target distributions.

\begin{figure*}[ht]
\centering
\includegraphics[width=0.60\linewidth]{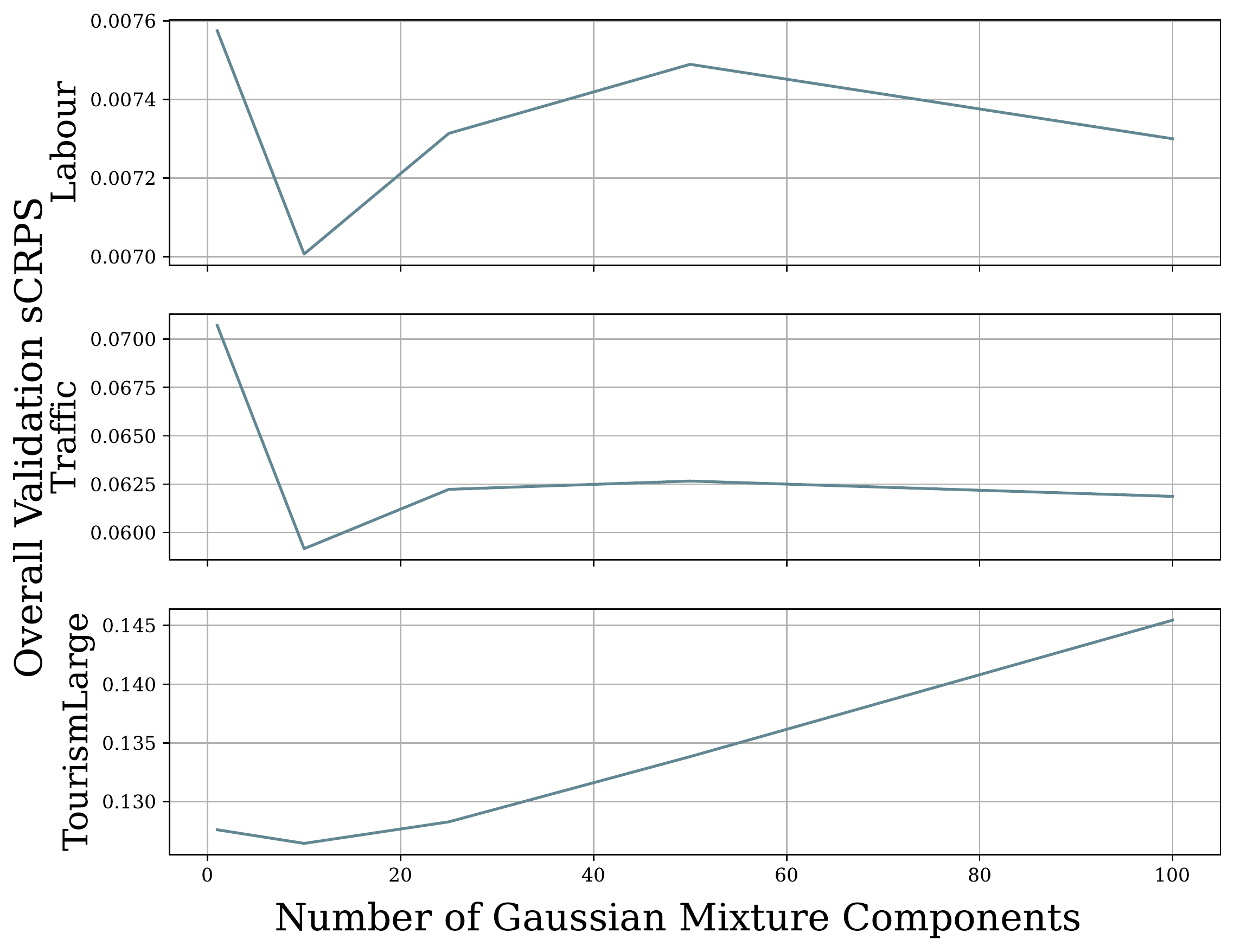}
\caption{
Validation \emph{scaled Continuous Ranked Probability Score} (sCRPS) for \Labour, \Traffic, and \TourismL. We show performance curves for \NHITS\ as a function of the number of Mixture Components. We observe a bias-variance tradeoff, where initially, the sCRPS decreases as the number of components increases and reaches an optimal value at K=10 components. From there, after that, we see the sCRPS worsening, thus giving us the classic U-shaped tradeoff pattern.} \label{fig:ablation_study_crps_vs_k}
\end{figure*}

\clearpage
\subsection{Architectures and Reconciliation Ablation Study}
\label{section:reconciliation_ablation}

As mentioned in Section~\ref{section:multivariate_mixture} and proven in Appendix \ref{section:hierarchical_mixture_properties}, the flexibility of the multivariate mixture probability model is compatible with most probabilistic reconciliation techniques. In this ablation study, we compare the accuracy effect of different reconciliation strategies. The reconciliation strategies considered are \BottomUp~\citep{orcutt1968hierarchical_bottom_up, dunn1976hierarchical_bottom_up2}, \TopDown~\citep{gross1990hierarchical_top_down, fliedner1999hierarchical_top_down2}, and \MinTrace~\citep{hyndman2011optimal_combination_hierarchical, wickramasuriya2019hierarchical_mint_reconciliation} variants. We describe them in detail below.

Consider the base forecasts $\hat{\mathbf{y}}_{[i],\tau} \in \mathbb{R}^{N_{a}+N_{b}}$, a reconciliation process uses a matrix $\mathbf{P}_{[b][i]}  \in \mathbb{R}^{N_{b} \times (N_{a}+N_{b})}$ that collapses the original base forecasts into bottom-level forecast that are later aggregated for the upper levels of the hierarchy into the
reconciled forecasts $\tilde{\mathbf{y}}_{[i],\tau}$. Here we use the convenient representation of the reconciliation strategies introduced in Section~\ref{section2:literature}.
\begin{equation*}
\tilde{\mathbf{y}}_{[i],\tau} = \mathbf{S}_{[i][b]} \mathbf{P}_{[b][i]} \hat{\mathbf{y}}_{[i],\tau}
= \mathbf{SP}\left( \hat{\mathbf{y}}_{[i],\tau} \right)
\end{equation*}

\textbf{Bottom-Up.}
The most basic hierarchical reconciliation consists of simply aggregating the bottom-level base forecasts $\hat{\mathbf{y}}_{[b],\tau}$. By construction, it satisfies the hierarchical aggregation constraints. 
\begin{equation}
    \mathbf{P}_{[b][i]}=[\mathbf{0}_{\mathrm{[b][a]}}\;|\;\mathbf{I}_{\mathrm{[b][b]}}]
\end{equation}

\textbf{Top-Down.}
The \TopDown\ strategy distributes an aggregate level forecast into the bottom-level forecasts using proportions $\mathbf{p}_{[b]}$. Proportions can be historical values, or they can be forecasted. 
\begin{equation}
    \mathbf{P}_{[b][i]}=[\mathbf{p}_{\mathrm{[b]}}\;|\;\mathbf{0}_{\mathrm{[b][a,b\;-1]}}]
\end{equation}

\textbf{MinTrace.}
Newer reconciliation strategies use all the information available throughout the hierarchy optimally. In particular, the \MinTrace\ reconciliation is proven to be the optimizer of a mean squares error objective that transforms base predictions into hierarchically coherent predictions under an unbiasedness assumption. Its reconciliation matrix is given by:
\begin{equation}
    \mathbf{P}_{[b][i]} =
    \left(\mathbf{S}^{\intercal}\hat{\boldsymbol{\Sigma}}_{\tau}\mathbf{S}\right)^{-1}
    \mathbf{S}^{\intercal}\hat{\boldsymbol{\Sigma}}^{-1}_{\tau}
\end{equation}

We summarize the ablation study results for the different reconciliation strategies in Table~\ref{table:ablation_reconciliation}. We report the overall sCRPS across five datasets for the \NHITS, \TCN, and \TFT\ architectures. We obtain the probabilistic predictions using bootstrap ~\citep{panagiotelis2023probabilistic_reconciliation}. We observe clear advantages from adopting novel reconciliation techniques such as \MinTrace\ as it improves accuracy over \BottomUp\ by 20 to 30 percent margins across well-established neural forecast architectures. 
We find that post processing reconciliation is capable of improving complex end-to-end approaches that integrate the hierarchical constraints into the training procedure \citealt{rangapuram2021hierarchical_e2e,kamarthi2022profhit_network,han2021hierarchical_sharq}.
Based on the results of these ablation studies, we conducted the main experiments of this work with the \MinTrace\ and the \BottomUp\ reconciliation techniques. 

\begin{table*}[htp] 
\tiny
\centering
\caption{
Empirical evaluation of probabilistic coherent forecasts. Mean scaled continuous ranked probability score (sCRPS), averaged over 10 random seeds, at each aggregation level. The best result is highlighted (lower measurements are preferred).\\
\tiny{
\textsuperscript{*} The \TopDown\ reconciliation is only available for strictly hierarchical datasets.
}
}
\label{table:ablation_reconciliation}
\setlength\tabcolsep{2.5pt}
\begin{tabular}{cl|cccccc}
\toprule
    & \textsc{Dataset}                    & \Base      & \MinTrace-ols      & \MinTrace-wls      & \TopDown-ap\textsuperscript{*}        & \TopDown-pa\textsuperscript{*}      & \BottomUp       \\
\parbox[t]{.2mm}{\multirow{5}{*}{\rotatebox[origin=c]{90}{\NHITS}}}
    &                            \Labour  &    0.0082  &   0.0082±0.0001    &   0.0084±0.0001    &   0.0092±0.0001    &   0.0091±0.0000  &  0.0094±0.0001  \\
    &                            \Traffic &    0.0629  &   0.0635±0.0011    &   0.0643±0.0010    &   0.0651±0.0010    &   0.0650±0.0013  &  0.0660±0.0008  \\
    &                            \Tourism &    0.0791  &   0.0806±0.0011    &   0.0771±0.0014    &   0.0920±0.0010    &   0.0913±0.0014  &  0.0756±0.0012  \\
    &                           \TourismL &    0.1274  &   0.1281±0.0004    &   0.1261±0.0006    &   -                &  -               &  0.1351±0.0005  \\
    &                           \Wikitwo  &    1.4531  &   1.3165±0.0302    &   1.8399±0.0904    &   0.5165±0.0159    &   0.5178±0.0088  &  3.3351±0.1690  \\ \midrule
\parbox[t]{.2mm}{\multirow{5}{*}{\rotatebox[origin=c]{90}{\TCN}}}
    &                            \Labour  &    0.0213  &   0.0243±0.0004    &   0.0202±0.0003    &   0.0237±0.0003    &   0.0237±0.0004  &  0.0187±0.0004  \\
    &                            \Traffic &    0.0566  &   0.0569±0.0007    &   0.0577±0.0007    &   0.0605±0.0010    &   0.0605±0.0009  &  0.0623±0.0008  \\
    &                            \Tourism &    0.0664  &   0.0660±0.0009    &   0.0641±0.0011    &   0.0803±0.0012    &   0.0807±0.0015  &  0.0678±0.0018  \\
    &                           \TourismL &    0.1632  &   0.1638±0.0010    &   0.1640±0.0008    &   -                &   -              &  0.1677±0.0007  \\
    &                           \Wikitwo  &    2.7345  &   2.0942±0.0539    &   2.9305±0.1247    &   1.6188±0.0374    &   1.6232±0.0476  &  4.2341±0.1338  \\ \midrule
\parbox[t]{.2mm}{\multirow{5}{*}{\rotatebox[origin=c]{90}{\TFT}}}    
    &                            \Labour  &    0.0073  &   0.0071±0.0001    &   0.0074±0.0000    &   0.0084±0.0001    &   0.0084±0.0001  &  0.0087±0.0001  \\
    &                            \Traffic &    0.0632  &   0.0641±0.0007    &   0.0638±0.0008    &   0.0650±0.0015    &   0.0646±0.0011  &  0.0658±0.0009  \\
    &                            \Tourism &    0.0944  &   0.1015±0.0010    &   0.0895±0.0007    &   0.0834±0.0008    &   0.0832±0.0006  &  0.0922±0.0013  \\
    &                           \TourismL &    0.1360  &   0.1364±0.0007    &   0.1362±0.0008    &   -                &   -              &  0.1442±0.0010  \\
    &                           \Wikitwo  &    0.2609  &   0.2608+0.0010    &   0.2641±0.0025    &   0.2560±0.0016    &   0.2560±0.0013  &  0.2755±0.0041  \\ \bottomrule
\end{tabular}
\end{table*}

\clearpage
\section{Software and Training Methodology}
\label{section:training_methodology}
\subsection{Hyperaparameters and Training Methodology}

\begin{table}[!ht] 
\caption{\ours\ fixed hyperparameters.}
\label{table:fixed_hyperparameters}
\centering
\tiny
\begin{tabular}{lccc}
\toprule
\textsc{Hyperparameter}     &   \multicolumn{3}{c}{\textsc{Fixed Values}}               \\ \midrule
Architecture                &  \NHITS           &  \TFT             &  \TCN             \\
Activation                  &  ReLU             &  ReLU             &  ReLU             \\
Encoder units               &  256              &  256              &  256              \\
Encoder layers*             &  4                &  3                &  4                \\
Encoder type                &  MLP              &  LSTM             &  Conv1D           \\
Train Objective             &  Comp.Lik.        &  Comp.Lik.        &  Comp.Lik.        \\ \bottomrule
\end{tabular}
\end{table}

\begin{table}[!ht] 
\caption{\ours\ optimized hyperparameters.}
\label{table:hyperparameters}
\scriptsize
\centering
\tiny
    \begin{tabular}{ll}
    \toprule
    \textsc{Hyperparameter}                               & \textsc{Considered Values}                                          \\ \midrule
    Initial learning rate.                                & \{1e-3,5e-4,1e-4\}                                                  \\
    Number of learning rate decays.                       & \{None, 3\}                                                         \\
    Training steps.                                       & \{.5e3, 1e3, 1.5e3, 2e3, 2.5e3, 3e3\}                               \\
    Input size multiplier (L=m*H).                        & $m \in \{2,3,4\}$                                                   \\
    Reconciliation strategy.                              & \{BottomUp,                                                         \\
                                                          &\;MinTraceOLS, MinTraceWLS \}                                        \\ \bottomrule
    \end{tabular}
\end{table}

Training \ours\ and the benchmark models involves dividing the data into training, validation, early stopping, and test sets, as shown in Figure~\ref{fig:train_methodology}. The training set consists of the observations before the last two horizon windows; validation is the window between the train and test sets, with test being the last window. The model's performance on the validation set guides the exploration of the hyperparameter space (\HYPEROPT, \citealt{bergstra2011hyperopt}). During the recalibration phase, we retrain the models to incorporate new information before being tested.

We followed a standard two-stage approach for hyperparameter selection. In the first stage, based on validation ablation studies from Appendix~\ref{section:reconciliation_ablation}, we fixed the architecture and the probability distribution to be estimated; Table~\ref {table:fixed_hyperparameters} describes the hyperparameters. Then conditional on a good performing architecture, in the second stage, we optimized the training procedure of the architecture, optimally exploring the space defined in Table~\ref{table:hyperparameters} with \HYPEROPT. This approach allowed us to explore the hyperparameter space while keeping it computationally tractable. It also demonstrated the \ours's robustness, broad applicability, and potential to achieve high accuracy with only slight adjustments.

We train \ours\ to maximize the composite likelihood from Equation~(\ref{eqn:composite_likelihood}) using the \ADAM~\citep{kingma2014adam_method} stochastic gradient algorithm. An early stopping strategy ~\citep{yuan2007early_stopping} is employed to halt training if there is no improvement in the overall sCRPS metric on the validation set.

\subsection{Software Implementation}

All statistical baselines use \StatsForecast's \AutoARIMA~\citep{ hyndman2008automatic_arima,garza2022statsforecast} and \HierarchicalForecast's reconciliation methods implementations~\citep{olivares2022hierarchicalforecast}. We created a unified Python re implementation of various widely-used hierarchical forecasting techniques that we make publicly available in the \HierarchicalForecast\ library~\citep{olivares2022hierarchicalforecast}. The shared implementation allows us to standardize the comparison of the methods, controlling for experimental details and guaranteeing the quality of statistical baselines. The code is publicly available in a dedicated repository to support reproducibility and related research.

Regarding the hierarchical neural forecast baselines, \HierEtoE~\citep{rangapuram2021hierarchical_e2e} is available in the \href{https://github.com/rshyamsundar/gluonts-hierarchical-ICML-2021}{\textcolor{blue}{GluonTS \ library}}, while \PROFHIT~\citep{kamarthi2022profhit_network}\ is available in a \href{https://github.com/AdityaLab/PROFHiT}{\textcolor{blue}{\PROFHIT\ dedicated repository}}. As mentioned earlier the only available implementation for \PROFHIT\ suffers from significant numerical instability in its optimization. We use the optimal configurations reported in \HierEtoE\ and \PROFHIT\ repositories.

The \ours\ framework is implemented in \PyTorch~\citep{pytorch2019library} and can be run on both CPUs and GPUs. We have made the \ours\ source code available, along with all the experiments in the following \href{https://anonymous.4open.science/r/HINT-FBA8/README.md}{\textcolor{blue}{\ours\ dedicated repository}}.

\clearpage
\section{Extended Main Results}
\label{section:extended_main_results}
Table~\ref{table:summarized_crps_evaluation} summarizes probabilistic and point forecast accuracy. We provide measurements of the best results between \NHITS, \TFT\ and a simple \TCN\ for different hierarchy levels, following the training and hyperparameter selection methodologies in Appendix~\ref{section:training_methodology}. The top row of each panel reports the overall sCRPS or relMSE. Moving from aggregate to disaggregate levels increases forecast errors. \ours\ outperforms the second-best alternative overall sCRPS by an average of 8.14\% across datasets, and 13.2\% without \Traffic. We observed improvements of \CRPSgainsLabour\ on \Labour, \CRPSgainsTourismS\ on \Tourism, \CRPSgainsTourismL\ on \TourismL, and \CRPSgainsWikitwo\ on \Wikitwo. However, for \Traffic, \HierEtoE\ outperforms \ours\ by \CRPSgainsTraffic\ due to the clear Granger causalities in the dataset, which merit the \HierEtoE's VAR approach. The sCRPS and relMSE results are highly correlated, although relMSE is less robust to outliers.

\begin{table*}[htp] 
\tiny
\centering
\caption{
Empirical evaluation of probabilistic coherent forecasts. Mean scaled continuous ranked probability score (sCRPS) and mean relative squared error (relMSE), averaged over 10 random seeds, at each aggregation level. 
The best result is highlighted.\tiny{
\textsuperscript{\dag} The \PROFHIT\ results differ from \cite{kamarthi2022profhit_network}, as the \href{https://github.com/AdityaLab/PROFHiT}{\textcolor{blue}{only available implementation}} suffers from significant numerical instability in its optimization. \textsuperscript{*} Best performing variant of \TopDown\ (avg. proportions, proportions avg.), and \MinTrace\ (ols, wls, shrinkage) reported. \textsuperscript{**} The \PERMBU/\TopDown\ only available for strictly hierarchical datasets.
}
}
\label{table:crps_evaluation}
\setlength\tabcolsep{2.pt}
\begin{tabular}{ccl|c|cc|ccc|ccc}
\toprule
    &                   &                     &       \ours          &     \multicolumn{2}{c|}{\OTHER} &      \multicolumn{3}{c|}{\BOOTSTRAP} &         \multicolumn{3}{c}{\PERMBU\textsuperscript{**}} \\
    & \textsc{Dataset}  &    \textsc{Level}   &  (Ours)                   &    \HierEtoE           &      \PROFHIT\textsuperscript{\dag}    &    \BottomUp &   \TopDown\textsuperscript{*}  &  \MinTrace\textsuperscript{*}  &  \BottomUp &  \TopDown\textsuperscript{*}  &          \MinTrace\textsuperscript{*}  \\ \midrule
\parbox[t]{.2mm}{\multirow{30}{*}{\rotatebox[origin=c]{90}{sCRPS}}}
    &    \parbox[t]{.2mm}{\multirow{5}{*}{\rotatebox[origin=c]{90}{\Labour}}}
         &                            Overall &   \textbf{.0067±.0000}    &        .0171±.0003     &        .2138±.0007   &  .0078±.0001    &  .0668±.0000    &  .0073±.0000          &  .0077±.0001 &  .0623±.0001 &  .0069±.0001 \\
    &    &                            Country &   \textbf{.0017±.0000}    &        .0052±.0003     &        .2097±.0038   &  .0021±.0001    &  .0012±.0001    &  .0013±.0001          &  .0026±.0001 &  .0014±.0001 &  .0016±.0001 \\
    &    &                             Region &   \textbf{.0047±.0001}    &        .0181±.0003     &        .2150±.0028   &  .0058±.0001    &  .0458±.0000    &  .0045±.0001          &  .0060±.0001 &  .0420±.0001 &  .0045±.0001 \\
    &    &                      Region/Gender &   \textbf{.0075±.0000}    &        .0188±.0003     &        .2161±.0012   &  .0088±.0001    &  .0798±.0001    &  .0087±.0001          &  .0084±.0001 &  .0739±.0001 &  .0078±.0001 \\
    &    &                      Reg/Gndr/Empl &   \textbf{.0128±.0000}    &        .0262±.0004     &        .2142±.0027   &  .0145±.0001    &  .1403±.0000    &  .0148±.0001          &  .0137±.0001 &  .1320±.0002 &  .0137±.0001 \\ \cline{3-12}
    &    \parbox[t]{.2mm}{\multirow{5}{*}{\rotatebox[origin=c]{90}{\Traffic}}}
         &                            Overall &   .0589±.0004             &  \textbf{.0426±.0008}  &        .1137±.0022   &  .0736±.0024    &  .0741±.0012    &  .0608±.0014          &  .0849±.0009 &  .0708±.0008 &  .0651±.0008 \\
    &    &                             Level1 &   .0340±.0008             &  \textbf{.0276±.0011}  &        .0899±.0059   &  .0468±.0031    &  .0301±.0020    &  .0299±.0020          &  .0651±.0012 &  .0373±.0011 &  .0367±.0011 \\
    &    &                             Level2 &   .0347±.0006             &  \textbf{.0287±.0009}  &        .0879±.0034   &  .0483±.0030    &  .0329±.0017    &  .0323±.0017          &  .0622±.0013 &  .0367±.0009 &  .0357±.0008 \\
    &    &                             Level3 &   .0392±.0005             &  \textbf{.0297±.0009}  &        .0926±.0032   &  .0530±.0025    &  .0360±.0013    &  .0385±.0014          &  .0614±.0010 &  .0383±.0009 &  .0405±.0010 \\
    &    &                             Level4 &   .1275±.0002             &  \textbf{.0845±.0003}  &        .1842±.0014   &  .1463±.0017    &  .1975±.0017    &  .1424±.0015          &  .1507±.0004 &  .1709±.0010 &  .1473±.0004 \\ \cline{3-12}
    &    \parbox[t]{.2mm}{\multirow{5}{*}{\rotatebox[origin=c]{90}{\Tourism}}}
         &                            Overall &  \textbf{.0536±.0004}     &        .0761±.0007     &        .1358±.0033   &  .0682±.0018    &  .1040±.0014    &  .0703±.0017          &  .0649±.0016 &  .0898±.0012 &  .0680±.0016 \\
    &    &                            Country &  \textbf{.0147±.0004}     &        .0400±.0009     &        .0941±.0151   &  .0290±.0028    &  .0333±.0025    &  .0335±.0026          &  .0267±.0023 &  .0329±.0021 &  .0333±.0025 \\
    &    &                            Purpose &  \textbf{.0360±.0004}     &        .0609±.0012     &        .1300±.0069   &  .0490±.0027    &  .0782±.0017    &  .0507±.0023          &  .0450±.0017 &  .0697±.0021 &  .0497±.0018 \\
    &    &                      State/Purpose &  \textbf{.0709±.0006}     &        .0914±.0008     &        .1323±.0076   &  .0828±.0016    &  .1399±.0010    &  .0845±.0016          &  .0793±.0014 &  .1176±.0013 &  .0806±.0014 \\
    &    &                     Region/Purpose &  \textbf{.0929±.0006}     &        .1122±.0007     &        .1867±.0031   &  .1118±.0012    &  .1646±.0010    &  .1124±.0013          &  .1087±.0017 &  .1390±.0014 &  .1085±.0016 \\ \cline{3-12}
    &    \parbox[t]{.2mm}{\multirow{9}{*}{\rotatebox[origin=c]{90}{\TourismL}}}
         &                            Overall &   \textbf{.1176±.0002}    &        .1424±.0019     &        .2139±.0014   &  .1375±.0013    &  -              &  .1313±.0009          &  -            &  -            &  -         \\
    &    &                            Country &   \textbf{.0325±.0006}    &        .0698±.0029     &        .1353±.0090   &  .0622±.0026    &  -              &  .0471±.0018          &  -            &  -            &  -         \\
    &    &                              State &   \textbf{.0606±.0006}    &        .0936±.0019     &        .1610±.0020   &  .0820±.0019    &  -              &  .0723±.0011          &  -            &  -            &  -         \\
    &    &                               Zone &   \textbf{.1025±.0004}    &        .1260±.0017     &        .1893±.0034   &  .1207±.0010    &  -              &  .1143±.0007          &  -            &  -            &  -         \\
    &    &                             Region &   \textbf{.1457±.0003}    &        .1653±.0016     &        .2277±.0022   &  .1646±.0007    &  -              &  .1591±.0006          &  -            &  -            &  -         \\
    &    &                            Purpose &   \textbf{.0706±.0006}    &        .0996±.0028     &        .1845±.0071   &  .0788±.0018    &  -              &  .0723±.0014          &  -            &  -            &  -         \\
    &    &                      State/Purpose &   \textbf{.1088±.0003}    &        .1317±.0021     &        .2160±.0031   &  .1268±.0017    &  -              &  .1243±.0014          &  -            &  -            &  -         \\
    &    &                       Zone/Purpose &   \textbf{.1772±.0003}    &        .1926±.0015     &        .2679±.0019   &  .1949±.0010    &  -              &  .1919±.0008          &  -            &  -            &  -         \\
    &    &                     Region/Purpose &   \textbf{.2426±.0005}    &        .2606±.0017     &        .3296±.0010   &  .2698±.0008    &  -              &  .2694±.0006          &  -            &  -            &  -         \\ \cline{3-12}
    &    \parbox[t]{.2mm}{\multirow{6}{*}{\rotatebox[origin=c]{90}{\Wikitwo}}}
         &                            Overall &   \textbf{.2447±.0007}    &        .2592±.0031     &        .4009±.0028   &  .2894±.0038    &  .3231±.0037    &  .2808±.0035          &  .3920±.0044 &  .4269±.0036 &  .3821±.0049 \\
    &    &                              World &   \textbf{.1247±.0016}    &        .1007±.0046     &        .1244±.0085   &  .1796±.0069    &  .1777±.0084    &  .1793±.0067          &  .1777±.0125 &  .1945±.0109 &  .1801±.0123 \\
    &    &                            Country &   \textbf{.1805±.0011}    &        .1963±.0037     &        .2775±.0141   &  .2392±.0047    &  .2437±.0058    &  .2232±.0043          &  .2778±.0073 &  .3036±.0029 &  .2684±.0066 \\
    &    &                             Access &   \textbf{.2546±.0010}    &        .2784±.0038     &        .4405±.0034   &  .2966±.0032    &  .3379±.0026    &  .2781±.0028          &  .4196±.0059 &  .4621±.0071 &  .4006±.0059 \\
    &    &                              Agent &   \textbf{.2699±.0010}    &        .2900±.0043     &        .4526±.0084   &  .3036±.0033    &  .3427±.0026    &  .2855±.0029          &  .4255±.0058 &  .4669±.0070 &  .4073±.0060 \\
    &    &                              Topic &   \textbf{.3938±.0016}    &        .4307±.0039     &        .7094±.0109   &  .4282±.0038    &  .5134±.0037    &  .4379±.0029          &  .6595±.0060 &  .7074±.0049 &  .6540±.0058 \\ \bottomrule \bottomrule
\parbox[t]{.2mm}{\multirow{30}{*}{\rotatebox[origin=c]{90}{relMSE}}}
    &    \parbox[t]{.2mm}{\multirow{5}{*}{\rotatebox[origin=c]{90}{\Labour}}}
         &                            Overall &  .5802±.0131              &        .5667±.0265     & $6.774 \times 10^3$  &  .5382±.0000    &  16.8204±.0000  & \textbf{.3547±.0000}  &               &               &            \\
    &    &                            Country &  .1032±.0130              &        .1536±.0284     & $9.424 \times 10^2$  &  .2362±.0000    &  .0542±.0000    & \textbf{.0729±.0000}  &               &               &            \\
    &    &                             Region &  .6502±.0276              &        1.1486±.0413    & $6.635 \times 10^2$  &  .8281±.0000    &  14.6118±.0000  & \textbf{.3740±.0000}  &               &               &            \\
    &    &                      Region/Gender &  1.6870±.0241             &        1.1206±.0395    & $4.113 \times 10^2$  &  .9021±.0000    &  35.6038±.0000  & \textbf{.7519±.0000}  &               &               &            \\
    &    &                      Reg/Gndr/Empl &  1.8103±.0131             &        1.4491±.0361    & $1.664 \times 10^2$  &  .8069±.0000    &  5.6047±.0000   & \textbf{.8041±.0000}  &               &               &            \\ \cline{3-9}
    &    \parbox[t]{.2mm}{\multirow{5}{*}{\rotatebox[origin=c]{90}{\Traffic}}}
         &                            Overall &  .1212±.0051              & \textbf{.0340±.0051}   & .4536±.0224          &  .1394±.0000    &  .0614±.0000    & .0744±.0000           &               &              &             \\
    &    &                             Level1 &  .1004±.0057              & \textbf{.0253±.0057}   & .4591±.0413          &  .1296±.0000    &  .0491±.0000    & .0634±.0000           &               &              &             \\
    &    &                             Level2 &  .1156±.0051              & \textbf{.0302±.0050}   & .4291±.0336          &  .1342±.0000    &  .0625±.0000    & .0690±.0000           &               &              &             \\
    &    &                             Level3 &  .1602±.0041              & \textbf{.0529±.0038}   & .4612±.0240          &  .1582±.0000    &  .0730±.0000    & .0958±.0000           &               &              &             \\
    &    &                             Level4 &  .8893±.0042              & \textbf{.4206±.0048}   & .7709±.0081          &  .6457±.0000    &  .6525±.0000    & .6194±.0000           &               &              &             \\ \cline{3-9}
    &    \parbox[t]{.2mm}{\multirow{5}{*}{\rotatebox[origin=c]{90}{\Tourism}}}
         &                            Overall & \textbf{.0387±.0007}      &        .1471±.0046     & .9745±.0803          &  .1002±.0000    &  .1919±.0000    & .1235±.0000           &               &              &             \\
    &    &                            Country & \textbf{.0200±.0013}      &        .1821±.0094     & 1.2240±.1474         &  .0841±.0000    &  .1328±.0000    & .1233±.0000           &               &              &             \\
    &    &                            Purpose & \textbf{.0342±.0009}      &        .1038±.0040     & .8208±.0487          &  .0778±.0000    &  .1669±.0000    & .0957±.0000           &               &              &             \\
    &    &                      State/Purpose & \textbf{.0750±.0020}      &        .1550±.0032     & .8511±.0489          &  .1563±.0000    &  .3482±.0000    & .1620±.0000           &               &              &             \\
    &    &                     Region/Purpose & \textbf{.0928±.0015}      &        .1772±.0027     & .7227±.0942          &  .2000±.0000    &  .3628±.0000    & .2007±.0000           &               &              &             \\ \cline{3-9}
    &    \parbox[t]{.2mm}{\multirow{9}{*}{\rotatebox[origin=c]{90}{\TourismL}}}
         &                            Overall &  \textbf{.0577±.0009}     &        .2449±.0096     & 1.0401±.0296         &  .3070±.0000    &  -              & .1375±.0000           &               &              &             \\
    &    &                            Country &  \textbf{.0336±.0009}     &        .2918±.0194     & 1.3473±.1430         &  .4399±.0000    &  -              & .1268±.0000           &               &              &             \\
    &    &                              State &  \textbf{.0598±.0012}     &        .2850±.0107     & 1.0854±.0358         &  .3504±.0000    &  -              & .1564±.0000           &               &              &             \\
    &    &                               Zone &  \textbf{.1263±.0019}     &        .3620±.0087     & 1.0821±.0397         &  .3950±.0000    &  -              & .2664±.0000           &               &              &             \\
    &    &                             Region &  \textbf{.1777±.0024}     &        .3594±.0055     & .9447±.0300          &  .3996±.0000    &  -              & .3211±.0000           &               &              &             \\
    &    &                            Purpose &  \textbf{.0478±.0006}     &        .1581±.0086     & .8257±.0758          &  .1624±.0000    &  -              & .0759±.0000           &               &              &             \\
    &    &                      State/Purpose &  \textbf{.0752±.0009}     &        .1884±.0055     & .7839±.0491          &  .1860±.0000    &  -              & .1332±.0000           &               &              &             \\
    &    &                       Zone/Purpose &  \textbf{.1638±.0017}     &        .2872±.0049     & .8178±.0187          &  .2932±.0000    &  -              & .2550±.0000           &               &              &             \\
    &    &                     Region/Purpose &  \textbf{.2296±.0016}     &        .3360±.0047     & .8013±.0215          &  .3661±.0000    &  -              & .3464±.0000           &               &              &             \\ \cline{3-9}
    &    \parbox[t]{.2mm}{\multirow{6}{*}{\rotatebox[origin=c]{90}{\Wikitwo}}}
         &                            Overall &   \textbf{.1884±.0012}    &        .6598±.0249     & .7901±.0384          &  1.0163±.0000   &  1.4482±.0000   & 1.0068±.0000          &               &              &             \\
    &    &                              World &   \textbf{.1955±.0037}    &        .2738±.0301     & .3274±.0798          &  .9245±.0000    &  1.6135±.0000   & .9883±.0000           &               &              &             \\
    &    &                            Country &   \textbf{.1648±.0015}    &        .7427±.0430     & .9150±.0886          &  1.0204±.0000   &  1.3529±.0000   & 1.0252±.0000          &               &              &             \\
    &    &                             Access &   \textbf{.1921±.0012}    &        .9575±.0331     & 1.1538±.0429         &  1.1267±.0000   &  1.4159±.0000   & 1.0267±.0000          &               &              &             \\
    &    &                              Agent &   \textbf{.1972±.0010}    &        .9384±.0381     & 1.1312±.0695         &  1.1008±.0000   &  1.3562±.0000   & 1.0047±.0000          &               &              &             \\
    &    &                              Topic &   \textbf{.1923±.0010}    &        1.0305±.0145    & 1.1881±.0492         &  1.0603±.0000   &  1.2282±.0000   & 1.0182±.0000          &               &              &             \\ \bottomrule    
\end{tabular}
\end{table*}

\end{document}